\documentclass[conference]{IEEEtran}
\usepackage{amsmath,amssymb,amsfonts}
\usepackage{graphicx}
\usepackage{textcomp}
\usepackage{xcolor}
\usepackage{booktabs}
\usepackage{algorithm}
\usepackage[switch]{lineno}
\usepackage{multirow}
\usepackage[hidelinks]{hyperref}
\usepackage{anyfontsize}
\usepackage{balance}
\usepackage{threeparttable}
\usepackage{float}
\usepackage{stfloats}
\usepackage{lipsum}
\usepackage{multicol}
\usepackage{float}
\usepackage{color}
\usepackage{amssymb}
\usepackage{bm}
\usepackage{subfigure}
\usepackage{algpseudocode}  
\usepackage[numbers,sort&compress]{natbib}
\def\BibTeX{{\rm B\kern-.05em{\sc i\kern-.025em b}\kern-.08em
    T\kern-.1667em\lower.7ex\hbox{E}\kern-.125emX}}
\begin{document}
\title{LIFE: Learning Individual Features for Multivariate Time Series Prediction with Missing Values}

% Single author syntax
\author{\IEEEauthorblockN{Zhao-Yu Zhang, Shao-Qun Zhang, Yuan Jiang, and Zhi-Hua Zhou}
\IEEEauthorblockA{National Key Laboratory for Novel Software Technology, Nanjing University, Nanjing 210023, China\\
Email: \{zhangzhaoyu, zhangsq, jiangy, zhouzh\}@lamda.nju.edu.cn    
}
}

\maketitle

\begin{abstract}
    Multivariate time series (MTS) prediction is ubiquitous in real-world fields, but MTS data often contains missing values. In recent years, there has been an increasing interest in using end-to-end models to handle MTS with missing values. To generate features for prediction, existing methods either merge all input dimensions of MTS or tackle each input dimension independently. However, both approaches are hard to perform well because the former usually produce many unreliable features and the latter lacks correlated information. In this paper, we propose a Learning Individual Features (LIFE) framework, which provides a new paradigm for MTS prediction with missing values. LIFE generates reliable features for prediction by using the correlated dimensions as auxiliary information and suppressing the interference from uncorrelated dimensions with missing values. Experiments on three real-world data sets verify the superiority of LIFE to existing state-of-the-art models.
\end{abstract} 
\begin{IEEEkeywords}
    Multivariate Time Series Prediction, Missing Values, Correlated Dimensions, Individual Features
\end{IEEEkeywords}

\setcounter{secnumdepth}{2}
\section{Introduction}
\label{sec:1}
Multivariate time series (MTS) data is prevalent in many fields, such as health care~\cite{song2018attend,shi2020iddsam}, weather forecasting~\cite{cao2018brits}, transportation prediction~\cite{tang2020joint}, quantitative trading~\citep{zhang2020hrp}, etc. However, real-world MTS data usually contains abundant missing values due to cost control, sensor damage, irregular sampling, and other reasons. Missing values, without a doubt, make it more difficult for analysts and engineers to deal with the tasks of time series analysis.

For MTS prediction with missing values, a natural idea is the two-step approach: first fill the missing values with replacement values, that is, \emph{data imputation}, and then apply the complete data to prediction. However, previous studies suggest that the separation of imputation and prediction processes has a proclivity for suboptimal results~\cite{mrnn2017,cao2018brits,tang2020joint}. Besides, it is usually unnecessary to perform imputation since many real-world tasks care more about prediction results than restoring original data. Recent years have witnessed an increasing interest in exploring end-to-end models, which perform better than the two-step methods~\cite{che2018recurrent,song2018attend,narayan2020survey}. These models usually estimate the missing values or treat them as zeros, and then merge all input dimensions~\cite{cao2018brits,tan2020datagru} or feed each dimension independently~\cite{zhang2019modelling,mimic3preprocess} into Recurrent Neural Networks (RNNs) or attention-based models to generate features for prediction.

\begin{figure}[ht]
    \includegraphics[width=.95\columnwidth]{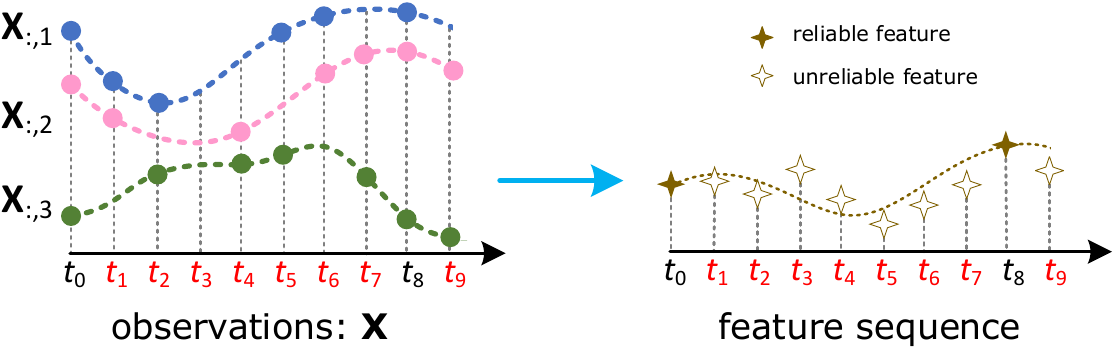}
    \centering
    \caption{Feature generation of most existing end-to-end models. Merging all input dimensions leads to numerous unreliable features.}
    \label{fig:rnn_models}
\end{figure}

However, merging all input dimensions may result in many unreliable features, which is caused by the interference of missing values. Fig.~\ref{fig:rnn_models} gives a vivid illustration of the feature generation process above, where $\mathbf{X}_{:,d}$ denotes the $d$-th dimension. At each timestamp, three input points are converted to a feature. Once there is a missing value of any input dimension, the generated feature is unreliable. Adhering to this line of thought, we can calculate that 30\% of input values are missing, but 80\% of generated features are unreliable. On the other hand, some approaches tackle each input dimension independently. Thus, the generated feature sequence has the same unreliable rate as the missing rate of inputs. Nevertheless, this manner ignores the information provided by the correlated dimensions. Take the inputs in Fig.~\ref{fig:rnn_models} as an example, it's observed $\mathbf{X}_{:,1}$ and $\mathbf{X}_{:,2}$ are highly correlated. If one dimension is missing and the other dimension is observed, we should employ the correlated dimensions as auxiliary information to make the generated feature sequence more credible.

\begin{figure*}[ht]
    \includegraphics[width=.9\textwidth]{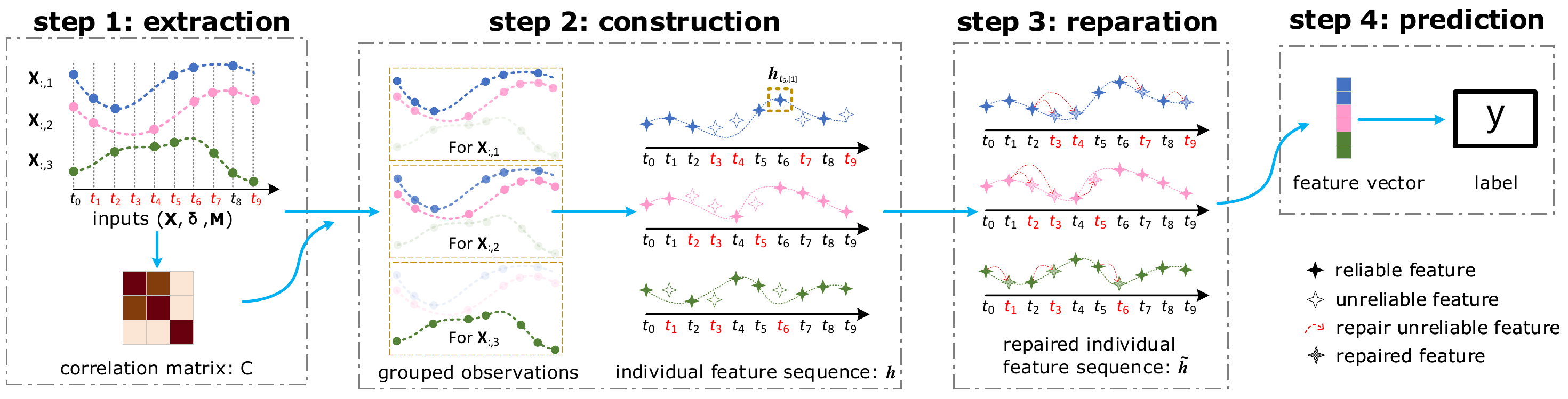}
    \centering
    \caption{Roadmap of LIFE framework. Steps 1 -- 4 corresponds to subsections~\ref{sec:CME} --~\ref{sec:prediction}, respectively.}
    \label{fig:roadmap}
\end{figure*}

In this paper, we propose a Learning Individual Features (LIFE) framework for MTS prediction with missing values. The roadmap is shown in Fig.~\ref{fig:roadmap}. The key idea of LIFE is Step 1 and Step 2, which collect \emph{credible and correlated} dimensions to build features, that is, \emph{individual features}. Individual features not only fuse with auxiliary information provided by correlated dimensions but also discard most of the dimensions uncorrelated to the concerned dimension. Therefore, LIFE can generate more reliable features than existing approaches. Our main contributions are summarized as follows:
\begin{itemize}
    \item We propose a novel framework LIFE, which provides a new paradigm for MTS prediction with missing values. LIFE builds individual features by credible and correlated dimensions, enabling it to generate many reliable features for the downstream prediction task.
    \item We present a general approach for extracting stable and credible dimensional correlations for MTS with missing values. We also provide two concrete algorithms for implementing this approach.
    \item We empirically verify that LIFE outperforms the state-of-the-art (SOTA) models on three real-world data sets.
\end{itemize}

The rest of the paper is organized as follows. Section~\ref{sec:2} reviews the related work. Section~\ref{sec:3} introduces some useful notations. Section~\ref{sec:4} describes the proposed LIFE framework. Section~\ref{sec:5} evaluates the performance of LIFE. Section~\ref{sec:6} concludes the paper.

\section{Related Work}
\label{sec:2}
\textbf{MTS prediction with missing values:} Recently, some end-to-end methods based on deep learning have achieved good performance in MTS prediction with missing values. These methods can be roughly divided into two categories: 1) Models based on RNNs generate features recurrently. Furthermore, the feature of the current timestamp is repaired concerning that of the adjacent timestamps. GRU-D~\cite{che2018recurrent} develops a decay mechanism to capture the temporal correlations to repair missing information. BRITS~\cite{cao2018brits} applies the decay mechanism to the bidirectional Long Short Term Memory (Bi-LSTM)~\cite{graves2005blstm} model. FG-LSTM~\cite{zhang2019modelling} focuses on modeling the temporal dependency for a single input dimension and can be regarded as running an LSTM for each dimension independently; 2) Models based on Self Attention employ attention mechanism instead of recurrence to generate features. SAnD~\cite{song2018attend} applies the attention-based method to healthcare applications without repairing the missing information. However, when generating features from inputs, the above methods calculate each input dimension independently or fuse all dimensions. Both manners would damage the performance because of the lack of dimensional correlations or the problem of unreliable features.

There have been many studies exploring dimensional correlations for MTS data without missing values. The Pearson Correlation Coefficient and Mutual Information can express the correlation between two time series. Besides, time series distance/dissimilarity can be used to imply correlation after a simple conversion (e.g., calculating the reciprocal or negative exponent). Some commonly used distances are as follows: Euler distance, Dynamic Time Warping (DTW)~\cite{berndt1994using}, and Optimal Transport / Wasserstein Distance~\cite{villani2008optimal,zhang2020time}. However, these methods may lead to fictitious correlations when the data contains missing values. Some kernel-based methods can be applied to MTS with missing values and provide similarity measurement~\cite{lu2008reproducing,cuturi2011fast,zhangsq2017}. Still, the missing values could also damage the results, especially with a large missing rate.

\section{Preliminaries}
\label{sec:3}

Let $\mathbf X \in \mathbb{R}^{T \times D}$ denote a $D$-dimensional time series with $T$ timestamps, and $\boldsymbol{s} = [s_t]_{t=1}^T \in \mathbb{R}^T$ is the corresponding timestamp sequence.
The masking matrix 
$
\mathbf M \in \{0,1\}^{T \times D}
$ indicates whether the values in $\mathbf X$ are missing:
\[
    {\mathrm M}_{td}=\left\{\begin{array}{ll}1, & \text { if } {\mathrm  X}_{td}  \text { is observed}; \\ 0, & \text { otherwise. }\end{array}\right.
\]
Let $\boldsymbol {{ \delta}}\in \mathbb{R}_+^{T\times D} $ denote the time interval matrix, which consists of the time gap ${\delta}_{td}$ from the timestamp of last observation to current timestamp. ${\delta}_{td}$ is defined as:
\[
    {\delta}_{td}=\left\{\begin{array}{ll}s_{t}-s_{t-1}+{\delta}_{t-1,d}, & t>1, {\mathrm  M}_{t-1,d}=0 ; \\ s_{t}-s_{t-1}, & t>1, {\mathrm  M}_{t-1,d}=1 ; \\ 0, & t=1.\end{array}\right.
\]

Note that subscripts $[\cdot]_{t,:}$ and $[\cdot]_{:,d}$ indicate the raw and column vectors of a matrix, respectively. For example, $\mathbf X_{t,:}$ and $\mathbf X_{:,d}$ denote the vectors of the $t$-th timestamp and $d$-th dimension, respectively. Suppose a vector $\boldsymbol{u} \in \mathbb{R}^{kD}$ consists of $D$ sub-vectors, each of which has a size of positive integer~$k$, then let $\boldsymbol{u}_{[d]}$ denote the $d$-th sub-vector:
\begin{equation}
    \label{eqn:sub_vector}
    \boldsymbol{u}_{[d]} = \left(u_{k(d-1)+1},~u_{k(d-1)+2},~ \dots~,~u_{kd} \right).
\end{equation}

Let the \emph{correlation matrix} $\mathbf C \in [0,1]^{D \times D}$ denote the pairwise correlations of $D$ dimensions in MTS, where ${\mathrm  C}_{ij}$ represents the correlation value between $\mathbf X_{:,i}$ and $\mathbf X_{:,j}$. The larger ${\mathrm  C}_{i j}$ is, the more relevant the two corresponding dimensions are. $\mathbf C$  is symmetric, and the diagonal elements are 1. For a special case that all dimensions are mutually independent, the correlation matrix $\mathbf C$ becomes an identity matrix.

This work aims to predict the supervised signals $\left\{y^{(n)}\right\}_{n=1}^N$ using the incomplete MTS values $\left\{\mathbf {X}^{(n)} \right\}_{n=1}^N$, time interval sequences $\left\{\boldsymbol \delta^{(n)}\right\}_{n=1}^N$, and masking matrices  $\left\{\mathbf {M}^{(n)} \right \}_{n=1}^N$, where $N$ denotes the number of MTS samples.

\section{LIFE Framework}
\label{sec:4}
In this section, we present the LIFE framework for MTS prediction with missing values. Fig.~\ref{fig:roadmap} shows the roadmap of LIFE. Step 1 extracts a credible and stable correlation matrix by penalizing missing values in qualifying dimensional correlation. In Step 2, we group the observations by the correlation matrix, and thus, build individual features using Self Attention. Step 3 repairs the individual features according to the temporal information of original data. In Step 4, we obtain the prediction results by the MTS classifiers or regressors. Finally, we can optimize the parameters in LIFE by jointly minimizing the imputation and prediction loss.

\subsection{Correlation Matrix Extraction}
\label{sec:CME}
This subsection aims to extract the credible and stable correlation matrix of the MTS data with missing values, which requires robustness to missing values and hyper-parameter settings. However, it is quite difficult for the conventional \emph{correlation matrix extraction} (CME) methods mentioned in Section~\ref{sec:2}. The reasons lie in two folds. First, some of these methods exclude the missing values and calculate similarity only when both points are observed. Obviously, this approach is not advisable because it discards much useful information. Second, the other methods first impute the missing values and then conduct CME. However, the imputation of missing values could be unreliable. Thus some dimensions with large missing rates may lead to unstable or even fictitious correlations. Experiments in Section~\ref{sec:cme_results} give some examples.

\begin{algorithm}[t]
    \caption{CME-PDTW algorithm}
    \label{alg:PDTW}
    \begin{algorithmic}[1]
        \Require 
        A data set with $N$ samples. The $n$-th sample consists of $D$-dimensional time series $\mathbf {X}^{(n)}$, time interval matrix $\boldsymbol{\delta}^{(n)}$ and masking matrix $\mathbf {M}^{(n)}$.
        \Ensure
        Correlation Matrix $\mathbf {C}$.
        \For{$n = 1 \to N$} \Comment{traverse each sample}
            \State $\mathbf {X}^{(n)} \!=\! \operatorname{interpolate}\!\left(\mathbf {X}^{(n)}\right)$ \Comment{impute missing values}
            \label{alg:PDTW:interpolation}
            \State $\mathbf {S}^{(n)} = \operatorname{zeros}(D,D)$  \Comment{PDTW distance}
            \State $\mathbf {Q}^{(n)} = \operatorname{zeros}(D,D)$  \Comment{weight of this sample}
            \For{$i = 1 \to D$}
            \label{alg:PDTW:startPDTWloop}
                \For{$j = i+1 \to D$}
                    \State $\mathrm {S}^{(n)}_{i j} = \mathrm {S}^{(n)}_{j i} = \operatorname{PDTW} \left(\mathbf{X}^{(n)}_{:,i}, \mathbf{X}^{(n)}_{:,j}\right)$
                    \label{alg:PDTW:dist}
                    \State ${\mathrm  Q}^{(n)}_{i j} = {\mathrm  Q}^{(n)}_{j i} = \operatorname{sum}\left(\mathbf{M}^{(n)}_{:,i} + \mathbf{M}^{(n)}_{:,j} \right)$ 
                    \label{alg:PDTW:weight}
                \EndFor
            \EndFor
            \label{alg:PDTW:endPDTWloop}
        \EndFor
        \State $\bar{\mathbf {S}} = \operatorname{weighted\_mean}(\mathbf {S}, \mathbf {Q})$
        \label{alg:PDTW:weighted_mean}
        \State $\mathbf {C}^{(\text{non-diag})} = \operatorname{normalize}\left( 1 / \bar{\mathbf {S}}^{(\text{non-diag})} \right) $
        \label{alg:PDTW:non-diag}
        \State $\mathbf {C}^{(\text{diag})} = 1 $
        \label{alg:PDTW:diag}
    \end{algorithmic}
\end{algorithm} 

To tackle the drawbacks above, we attempt to punish the missing values for CME. Since the imputation of missing values could be untrustworthy, we rely more on observed values and trust the correlations extracted by dimensions with low missing rates. Note that we are not trying to give each pairwise dimensional correlation a precise estimation but only look for those reliable and stable correlations. We believe penalizing missing values is a general method for CME and can be applied to various distances. Here, we give two concrete implementations: a DTW-based algorithm in the following and an Optimal-Transport-based algorithm in Appendix.

The key idea of Correlation Matrix Extraction - Penalty Dynamic Time Warping (CME-PDTW) algorithm is to convert the pairwise PDTW distances to a dimensional correlation matrix. It is worth mentioning that PDTW is not a legal distance measure, so it is more rigorous to call it ``PDTW dissimilarity". But for the sake of brevity, we use the term ``PDTW distance" without causing misunderstanding. The PDTW distance between $ \mathbf X_{:,d_1}$ and $\!\mathbf X_{:,d_2}  $ works by adding penalties of missing values to the original DTW algorithm:
\begin{equation}
    \label{eqn:pdtw}
    \resizebox{.91\columnwidth}{!}{ $ \displaystyle
        \!\!\!\!\operatorname{PDTW}(\mathbf X_{:,d_1}, \mathbf X_{:,d_2}\!)=\min_\pi \!\!\!\! \sum_{(i, j) \in \pi} \!\!\!\! \left[    \left(\mathrm  X_{i d_1} \!\!-\! \mathrm  X_{j d_2} \right)^2 \!\!+\! \phi \! \left(\mathrm  X_{i d_1}\!,\! \mathrm  X_{j d_2} \right) \right]$,
    }
\end{equation}
where $\pi$ is the search path of DTW and $\phi(\cdot)$ calculates the corresponding penalty term of missing values. The farther the last observation is, the more unreliable the estimation of missing value is. So $\phi(\cdot)$ is supposed to impose more punishment on the consecutive missing values than sporadic missing values. For $i,j$, we here formulate $\phi(\cdot)$ as follows:
\begin{equation}
	\label{eqn:pdtw_penalty}
	\resizebox{.91\columnwidth}{!}{ $ \displaystyle \!\! \!
		\phi \left(\mathrm  X_{i d_1}, \mathrm  X_{j d_2} \right) = p~\bigg[\mathrm \delta_{i d_1} (1\!-\!{\mathrm  M}_{i d_1}) + \mathrm \delta_{j d_2} (1\!-\!{\mathrm  M}_{j d_2})\bigg]
		$.
	}
\end{equation}

The produce of CME-PDTW is shown in Algorithm~\ref{alg:PDTW}. We use linear interpolation to estimate the missing values (Line~\ref{alg:PDTW:interpolation}) and calculate the pairwise PDTW distances for each sample (Lines~\ref{alg:PDTW:startPDTWloop}-\ref{alg:PDTW:endPDTWloop}). Next, we take the weighted mean over all samples to obtain the averaged distance matrix $\bar{\mathbf {S}}$, and the weight is proportional to the number of observations (Line~\ref{alg:PDTW:weighted_mean}). Thus, $\bar{\mathbf {S}}$ is mainly based on those samples with few missing values. Finally, we transform the distance matrix to the correlation matrix (Line~\ref{alg:PDTW:non-diag}-\ref{alg:PDTW:diag}). The interference of missing values is inhibited from two aspects to extract credible and stable correlations. On the one hand, samples with more missing values are assigned with smaller weights, which weakens their influence on the results. On the other hand, missing values will cause large PDTW distances, leading to small values in the correlation matrix. Thus, only those credible and stable correlations can be extracted, mainly relying on the samples and dimensions with low missing rates. 

CME-PDTW has two main advantages: 1) It can extract a credible and stable correlation matrix for MTS with missing values, even when the missing rate is extremely high; 2) The generated correlation matrix is pretty robust to the hyper-parameter penalty coefficient $p$, leading to the ease for adjusting this hyper-parameter. The experiments in Section~\ref{sec:5} verify the above advantages.

\begin{figure}[t]
    \includegraphics[width=0.75\columnwidth]{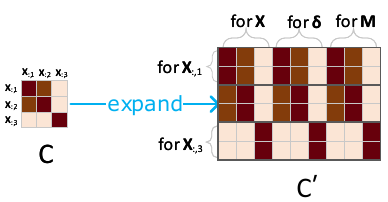}
    \centering
    \caption{Expansion of the correlation matrix $\mathbf{C}$. The size of each individual embedding vector is 2 $\left(k=2 \right)$ and the inputs size is three times of observations. So $\mathbf C$ ($3 \times 3$) is expanded to $\mathbf C^\prime$ ($6 \times 9$).}
    \label{fig:build_individual_feature}
\end{figure}
\subsection{Individual Features Construction}
The correlation matrix is used to collect the correlated dimensions for each concerned input dimension. Firstly, provided a positive integer $k$, which controls the model complexity, we expand $\mathbf  C$ to a larger matrix $\mathbf  C^\prime \in \mathbb{R}^{ kD \times 3D}$:
\[
	\label{eqn:expand}
	{\mathrm  C}^\prime_{i j}={\mathrm  C}_{i^\prime j^\prime}, \quad i^\prime = \lceil i/k\rceil , \quad \text{and}\quad j^\prime =  j  \bmod D,
\]
where $\lceil \rceil$ denotes the rounding up operation. See Fig.~\ref{fig:build_individual_feature} for an illustration of correlation matrix expansion.   Next, let $\mathbf{I}_t = (\mathbf X_{t,:}, \boldsymbol \delta_{t,:}, \mathbf M_{t,:} ) $ denote the input vector at time $t$ and any missing values are treated as zero.  We can transform $\mathbf{I}_t$ into an embedding vector $\boldsymbol{e}_t \in 
\mathbb{R}^{kD}$ by a one-layer neural network and the \emph{positional encoding} $\mathrm{pe}(\cdot)$ as following:
\begin{equation} \label{eq:e}
	\boldsymbol{e}_t\!=\!\sigma \left[ \mathbf{W} \!\odot\! \mathbf  C^\prime \times   \mathbf{I}_t  +  \boldsymbol{b} \right] +  \mathrm{pe}(t), 
\end{equation}
%\[
%	\boldsymbol{e}_t\!=\!f \left(\mathbf X_{t,:}, \boldsymbol \delta_{t,:}, \mathbf M_{t,:} ;~ \mathbf{W} \!\odot\! \mathbf  C^\prime, \boldsymbol{b} \right) + \mathrm{pe}(t), 
%\]
where $\mathbf{W} \in \mathbb{R}^{ kD \times 3D}$ is the connection weights,  $\boldsymbol{b} \in \mathbb{R}^{ kD}$ is the bias, $\odot$ and $\times$ are element-wise and matrix product, respectively, $\sigma$ is the sigmoid function, and $\mathrm{pe}(\cdot)$ is an embedding layer (a.k.a. lookup table) which maps the one-hot encoded timestamp $t$ to an $kD$-dimensional vector. Notice that the multiplier $\mathbf{W} \!\odot \mathbf  C^\prime$ only uses the correlated inputs for each concerned dimension, and thus, group the observations by correlated dimensions. In contrast with the conventional manners that merge all input dimensions, our approach can discard the uncorrelated information. As a result, the $k$-dimensional sub-vector ${\boldsymbol{e}}_{t,[d]} = ({e}_{t,k(d-1)+1},~ \dots~,~ {e}_{t,kd})$, as defined in Eq.~\ref{eqn:sub_vector}, only relies on the correlated observations corresponding to $\mathrm{X}_{td}$. We call this sub-vector $ {\boldsymbol{e}}_{t,[d]}$ the \emph{individual embedding vector} for $d \in \{1,2,\dots,D\}$. Thus, the embedding vector $\boldsymbol{e}_t$ consists of $D$ individual embedding vectors.

Provided the embedding vector $\boldsymbol{e}_{t}$, we can obtain the \emph{individual features} $\boldsymbol{h}_t$ using Self Attention. Formally, for each timestamp $t$, we use $\boldsymbol {e}_t $ as the query vector and generate the feature vector $\boldsymbol {h}_t$ as follows:
\[
\boldsymbol {h}_t = \sum_{s=1}^{T} a_s \boldsymbol {e}_{s} \quad\text{and}\quad a_s = \text{softmax}\left(f\left(\boldsymbol {e}_{s}, \boldsymbol {e}_{t} \right)\right) ,
\]
where $f$ is a one-layer network that calculates dependency score between $ \boldsymbol {e}_{s}$ and $\boldsymbol {e}_{t}$. Notice that the individual vector $\boldsymbol{h}_t $ still has the same size as $ \boldsymbol{e}_t $.  As shown in Fig.~\ref{fig:roadmap}, we can obtain the individual feature sequence $ [\boldsymbol{h}_t ]_{t=1}^T $ in Step 2, and an example of individual feature is $\boldsymbol{h}_{t_6,[1]}$, which is a $k$-dimensional vector corresponding to $\mathrm{X}_{t_6,1}$.

\subsection{Feature Reparation}
\label{sec:feature_reparation}

Note that some individual features $ [\boldsymbol{h}_t ]_{t=1}^T $ are still unreliable. If $\mathrm{X}_{td}$ is missing, then the individual feature $ {\boldsymbol{h}}_{t,[d]}$  is unreliable (e.g., the hollow star in Fig.~\ref{fig:roadmap}) according to Eq.~\eqref{eq:e}. A natural method is to repair the unreliable by temporal dependency. Here, we employ the decay mechanism~\cite{che2018recurrent} and obtain the repaired individual feature $\boldsymbol {\tilde{h}}_{t,[d]}$ as follows:
\[
	\boldsymbol {\tilde{h}}_{t,[d]}=\left\{\begin{aligned}
		&\boldsymbol {h}_{t,[d]}, & \text { if } {\mathbf  M}_{td} = 1; \\  & \gamma_{{t}d} ~ \boldsymbol {h}_{t^\prime\!,[d]} + \left(1- \boldsymbol \gamma_{{t}d}\right) \boldsymbol {h}_{t,[d]}, & \text { if } {\mathbf  M}_{td} = 0,
	\end{aligned}\right.
\]
where $t^\prime = t-{\boldsymbol \delta}_{td}$ is the timestamp of the last observed value for ${\mathrm  X}_{td}$, ${\boldsymbol \gamma_{{t}d}}$ is the decay rate, which indicates the how much information of last observed value remains. Intuitively, the decay rate can be calculated according to:  
\[
    \boldsymbol \gamma_{td}=\exp \left[-\max \left(0, \ w_d \boldsymbol \delta_{td} + a_d \right)\right],
\]
where both $\boldsymbol {w} \!=\! [w_d]_{d=1}^D \!\in\! \mathbb{R}^D$ and  $\boldsymbol {a} \!=\! [a_d]_{d=1}^D \!\in\! \mathbb{R}^D$ are learnable parameters. The repair process is illustrated in Fig.~\ref{fig:roadmap}, for example, both the hollow stars $\tilde{\boldsymbol{h}}_{t_3,[1]}$ and $\tilde{\boldsymbol{h}}_{t_4,[1]}$ are repaired by $\tilde{\boldsymbol{h}}_{t_2,[1]} = {\boldsymbol{h}}_{t_2,[1]}$.

\begin{algorithm}[t]
    \caption{Dense Interpolation for a sequence}
    \label{alg:dense}
    \begin{algorithmic}[1]
        \Require
        Feature sequence $[\tilde{\boldsymbol{h}}_{t}]_{t=1}^T$ with its length $T$ and dimension $D$, number of interpolation timestamps $F$.
        \Ensure
        Dense interpolated vector $\boldsymbol{v}$.
        \State $\mathbf {V}=\operatorname{zeros}(F,D)$
        \For{$t = 1 \to T$}
            \State $p=F*t/T$  
            \For{$f = 1 \to F$}
                \State $w=\operatorname{pow}\left(1-\operatorname{abs}\left(p-f\right) /F, 2\right)$   
                \State $\mathbf {V}\!_{f,:} = \mathbf {V}\!_{f,:}$ + $w \tilde{\boldsymbol{h}}_t$
            \EndFor
        \EndFor
        \State $\boldsymbol{v} = \operatorname{flatten}\left(\mathbf {V}\right)$
    \end{algorithmic}
\end{algorithm}

\subsection{Prediction}
\label{sec:prediction}

Provided the repaired individual  features $(\boldsymbol{\tilde{h}}_1, \boldsymbol{\tilde{h}}_2, \dots, \boldsymbol{\tilde{h}}_T)$, we can obtain the output using a MTS classifier or regressor. Notice that the size $T$ of repaired individual features may be different for each instance since the instance is irregular. Here, we unfold the classifier (or regressor) as follows: 
\[
	\hat{y} = f_{out} \left( f_{agg}\left( \boldsymbol{\tilde{h}}_1, \boldsymbol{\tilde{h}}_2, \dots, \boldsymbol{\tilde{h}}_T  \right)\right),  
\]
where $\hat{y}$ denotes the output, $f_{agg}$ aggregates the feature sequence $[ \boldsymbol{\tilde{h}}_t ]_{t=1}^T$ to a fixed-size vector, and $f_{out}$ is a one-hidden-layer network with softmax for classification or only linear layer for regression. We implement $f_{agg}$ by Dense Interpolation~\cite{trask2015modeling}, which shows better performance than the mean pooling and the attention pooling. The calculation process of Dense Interpolation is shown in Algorithm~\ref{alg:dense} and the key idea is to calculate the weighted mean of the whole sequence at specific timestamps.

LIFE jointly optimize both prediction and imputation loss:
\[
	{L}= {L}_{pred} + \alpha {L}_{imp},
\]
where ${L}_{pred}$ is the prediction loss, ${L}_{imp}$ denotes the imputation loss, and $\alpha \in \mathbb{R}_+$ is a balancing weight. The prediction loss relies on a specific task. For example, we usually employ the cross-entropy loss for a classification task and the mean square error (MSE) for a regression/forecasting task.  The intuition of adding imputation loss is to provide more supervised information for our model, leading to a better representation of the concerned MTS data. Here, we formulate ${L}_{imp}$ as the MSE between the observed and imputation values:
\[
	\resizebox{.87\columnwidth}{!}{ $ \displaystyle
		{L}_{imp} = \left. \sum_{t=1}^T \sum_{d=1}^D {\mathrm  M}_{td} \left({\mathrm  X}_{td}-{  \hat{\mathrm X}}_{td}\right)^2 \middle/ \sum_{t=1}^T \sum_{d=1}^D {\mathrm  M}_{td} \right. , $}
\]
where the imputation values $\hat{\mathrm X}_{td}$ are generated by 
\[
	{  \hat{\mathrm X}}_{td} = g\left(\boldsymbol {\tilde{h}}_{t,[d]}\right),
\]
in which $g(\cdot)$ is a one-layer perceptron.

\section{Experiments}
\label{sec:5}
In this section, we will evaluate the performance of the proposed model and the CME-PDTW algorithm. 
We are concerned about the following questions:
\begin{itemize}
    \item Does LIFE outperform the SOTA methods?
    \item Does LIFE benefit from learning individual features, and in what cases? What is the performance of LIFE with different correlation matrices?
    \item What is the impact of missing values on CME? Can CME-PDTW find credible and stable correlations?
\end{itemize}

\begin{table}[t]
    \centering
    \caption{attribute names and missing rates of human activity data.}
    \resizebox{\columnwidth}{!}{
        \begin{tabular}{c c c c c }
        \specialrule{0em}{1pt}{1pt}
        % \toprule
        \midrule
        \textbf{Index} &1 & 2 & 3 & 4 \\
        \textbf{Position}  & Ankle(L): x & Ankle(L): y & Ankle(L): z & Ankle(R): x \\ 
        \textbf{AMR} & 0.2012  & 0.2012 & 0.2012 & 0.2199 \\
        \midrule
        \textbf{Index} &5 & 6 & 7 & 8 \\
        \textbf{Position}  & Ankle(R): y & Ankle(R): z & Belt: x & Belt: y \\ 
        \textbf{AMR} & 0.2199  & 0.2199 & 0.2121 & 0.2121 \\
        \midrule
        \textbf{Index} &9 & 10 & 11 & 12\\
        \textbf{Position}  & Belt: z & Chest: x & Chest: y  & Chest: z \\ 
        \textbf{AMR} & 0.2121 &0.3445 & 0.3445& 0.3445 \\
        \midrule
        \end{tabular}
        }
        \begin{tablenotes}
            \item[1] ``AMR'' is short for ``average missing rate''. 
          \end{tablenotes}
    \label{tbl:human:missing_rate}
\end{table}

\subsection{Data Sets}
Here, we conduct the experiments on three real-world benchmark data sets widely used in community~\cite{narayan2020survey}. \textbf{PhysioNet}~\cite{silva2012predicting} data set contains 4000 records from intensive care unit (ICU). Following~\cite{cao2018brits}, we select 35 attributes that contain enough non-missing values and fix the sampling interval to 1 hour. Hence, each record has 48 timestamps containing 35 measurements (such as pH, heart rate, etc.). On this data set, we do mortality classification, which is a binary classification task. \textbf{Medical Information Mart for Intensive Care (MIMIC-\uppercase\expandafter{\romannumeral3})}~\cite{mimic3} data set with over 60,000 ICU stays is collected at Beth Israel Deaconess Medical Center from 2001 to 2012. After preprocessing introduced by~\cite{mimic3preprocess}, the resulting data has 42,276 ICU stays with 17 measurements (such as pH, temperature, etc.) and over 31 million clinical events. The first 48 hours of a stay are used as a sample. We conduct mortality classification and length-of-stay prediction (forecasting). \textbf{Human Activity}~\cite{kaluvza2010agent} is the UCI localization data, which contains 3D positions of the waist, chest, and ankles (12 dimensions in total) collected from five people performing 11 activities (e.g., walking, falling, sitting down). Each dimension was recorded for every 100 -- 200 milliseconds. We fix the sampling interval to 100 milliseconds and take the MTS records of 1000 milliseconds as one sample without overlap. There are 4,817 samples in total, and each MTS sample has 10 timestamps and 12 dimensions. The name and missing rates of 12 dimensions are shown in Table~\ref{tbl:human:missing_rate}. In real-world applications, sensor damage can cause high missing rates of corresponding dimensions. Instead of randomly dropping each observation with equal probability~\cite{cao2018brits}, we first randomly choose $n$ sensors (dimensions) as ``damaged sensors'', and then randomly eliminate 90\% observations of the damaged sensors. We alter $n$ from 0 -- 11 to generate 12 data sets. We believe the data generated by this method is much closer to real-world applications. Table~\ref{tbl:human:damaged_sensor} gives the damaged sensors and missing rates of the generated data sets in our experiments.  We do multi-class classification task on these data sets.

\begin{table}[t]
    \centering
    \caption{Details about the damaged sensors and average missing rates of human activity data set after preprocessing.}
    \resizebox{\columnwidth}{!}{
        {\large
        \begin{tabular}{c c c c c }
            \specialrule{0em}{1pt}{1pt}
            \toprule
            \textbf{$n$} & 0 &1 & 2 & 3 \\
            \textbf{$I$}  & $I_0=\{\}$ & $I_1=I_0 \cup \{7\}$ & $I_2=I_1 \cup \{12\}$ & $I_3=I_2 \cup \{5\}$  \\ 
            \textbf{AMR}  & 0.2445  & 0.3033 & 0.3525 & 0.4111 \\
            \toprule
            \textbf{$n$} & 4 &5 & 6 & 7  \\
            \textbf{$I$}  & $I_4=I_3 \cup \{11\}$ &  $I_5=I_4 \cup \{3\}$ & $I_6=I_5 \cup \{9\}$ & $I_7=I_6 \cup \{2\}$ \\ 
            \textbf{AMR} & 0.4605 & 0.5203 & 0.5795 & 0.6395 \\
            \toprule
            \textbf{$n$} & 8 & 9 & 10 & 11 \\
            \textbf{$I$}  & $I_8=I_7 \cup \{8\}$  & $I_9=I_8 \cup \{10\}$& $I_{10}=I_9 \cup \{4\}$ & $I_{11}=I_{10} \cup \{1\}$ \\ 
            \textbf{AMR} & 0.6985 &0.7475 & 0.8061& 0.8661 \\
            \bottomrule 
            \end{tabular}
            }

        }

        \label{tbl:human:damaged_sensor}
        \begin{tablenotes}
            \item[1] AMR is the average missing rate, $n$ is the number of damaged sensors, $I$ denotes the damaged sensor indexes, e.g., $I_3=\{7,12,5\}$ indicates that the sensors with index 7, 12, and 5 are damaged.  
          \end{tablenotes}
\end{table}

\subsection{Benchmark Methods}
We use two kinds of methods as baselines: (1) ``two-step" models, which comprise imputation and prediction independently, and (2) ``end-to-end" models, which jointly optimize both imputation and prediction processes.

\vspace{5pt}
\subsubsection{Two-Step Models}
\begin{itemize}
    \item \textbf{ROCKET-a} fills in missing values with the average observations over time, and then, uses ROCKET~\cite{rocket} for prediction, which is one of the SOTA classifiers for MTS without missing values.
    \item \textbf{ROCKET-m} employs MICE~\cite{azur2011multiple} for imputation and performs prediction via ROCKET as well as ROCKET-a.
    \item \textbf{LSTM-a} fills in missing values with the average observations over time, and then, uses LSTM for prediction.
    \item \textbf{LSTM-m} employs MICE for imputation and performs prediction via LSTM.
\end{itemize}

\subsubsection{End-to-End Models}
\begin{itemize}
    \item \textbf{GRU-D}~\cite{che2018recurrent} fuses Gated Recurrent Unit (GRU)~\cite{Kyunghyun2014GRU} with a decay mechanism for repairing information loss caused by missing values.
    \item \textbf{BRITS}~\cite{cao2018brits} establishes model based on bidirectional LSTM with the decay mechanism. 
    \item \textbf{SAnD}~\cite{song2018attend} is proposed for healthcare classification problem based on Self Attention. 
    \item \textbf{FG-LSTM}~\cite{zhang2019modelling} runs an LSTM for time-series dimension independently, and then, merges the features for prediction by a fully connected layer. 
\end{itemize}
Notice that GRU-D, BRITS, and SAnD merge all input dimensions, while FG-LSTM handles each dimension independently.

\subsection{Settings}

\begin{figure*}[t]
    \centering
    {\small
    \begin{minipage}[t]{.88\textwidth}{
        \resizebox{\textwidth}{!}{
            \centering{
                \subfigure[observation rate]{
                    \label{fig:correlation_matrix:missing_rate}
                    \centering
                    \includegraphics[width=0.2\textwidth]{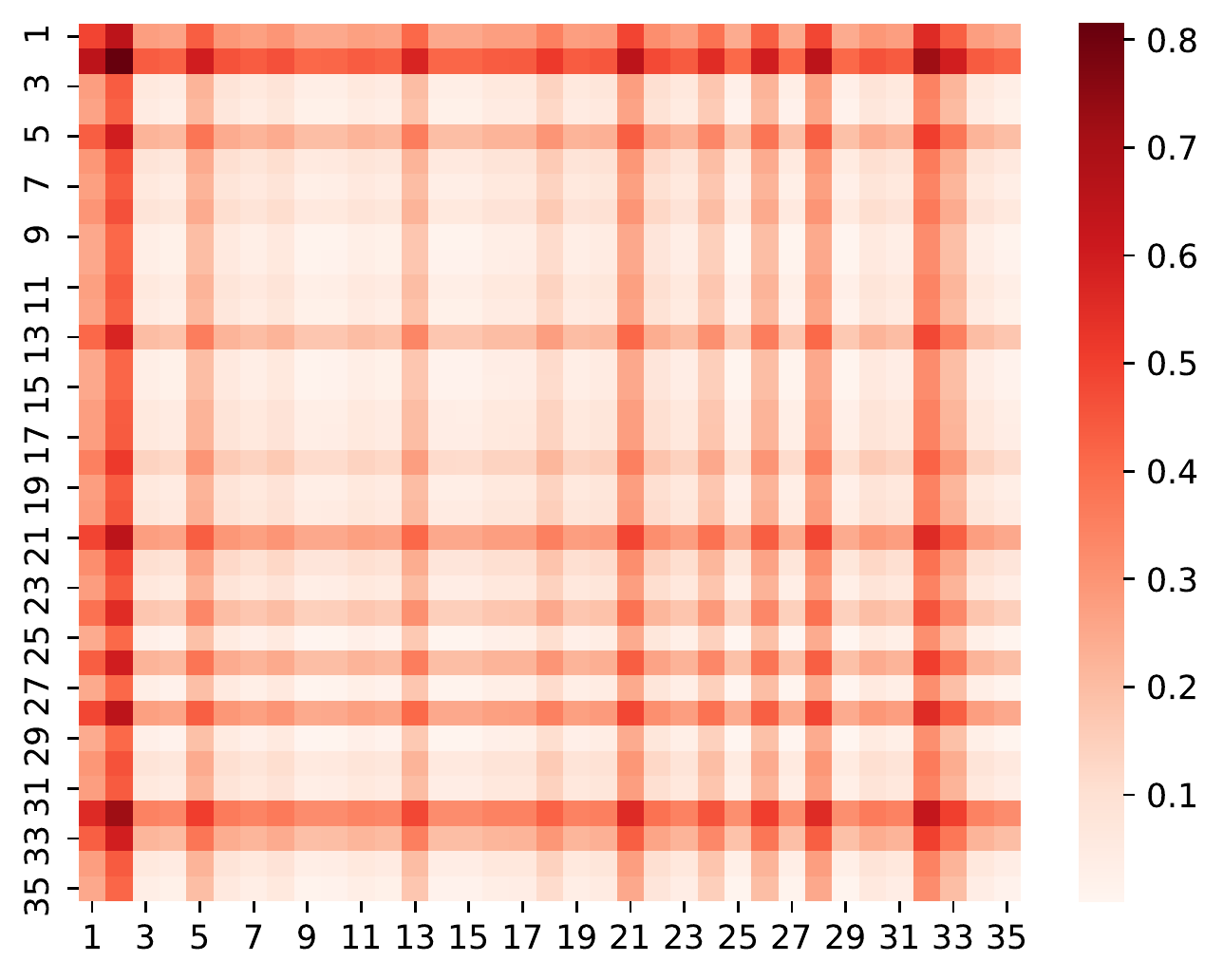}
                    }~~
                \subfigure[CME-Pearson]{
                    \centering
                    \includegraphics[width=0.2\textwidth]{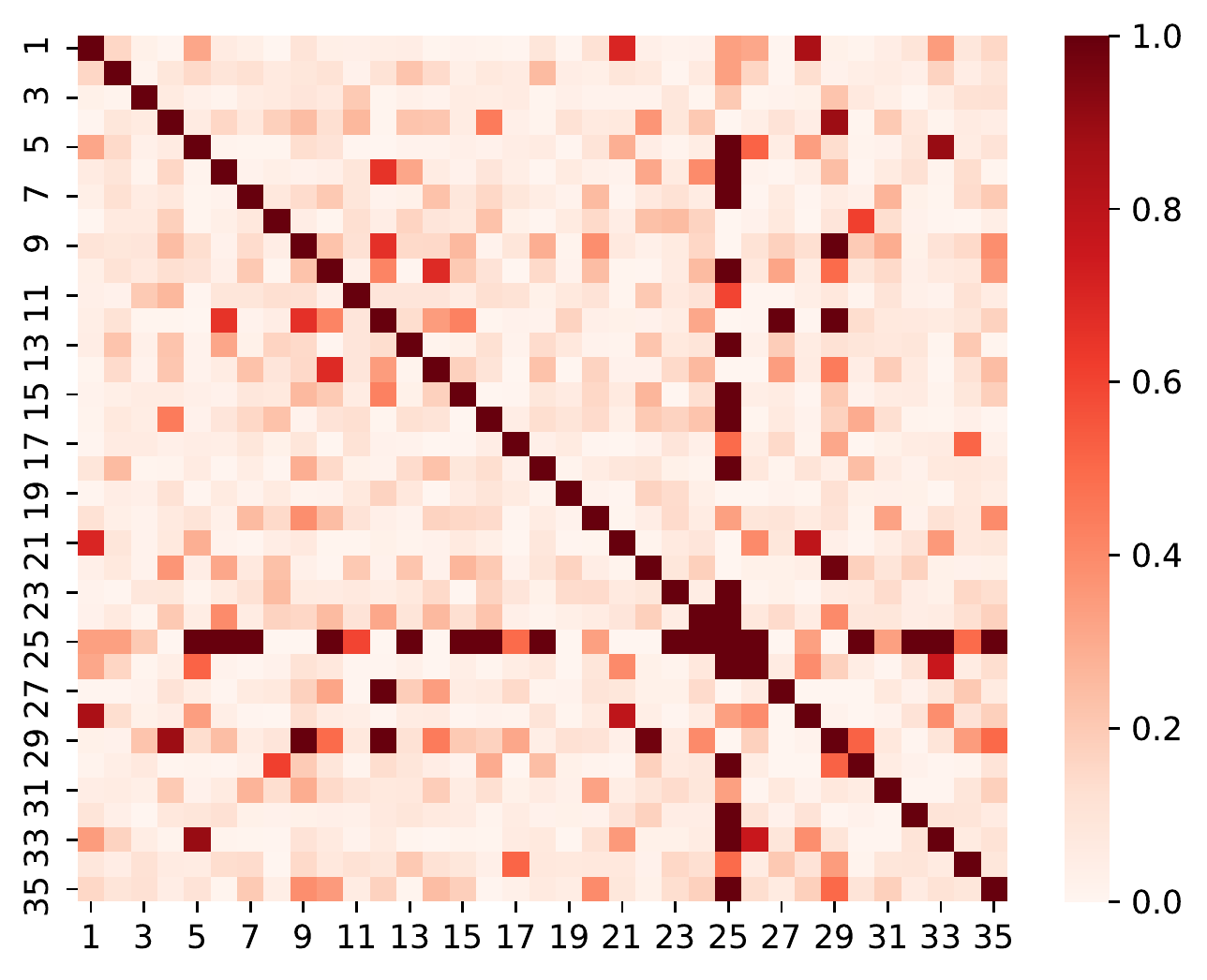}
                    }~~
                \subfigure[CME-DTW-i]{
                    \centering
                    \includegraphics[width=0.2\textwidth]{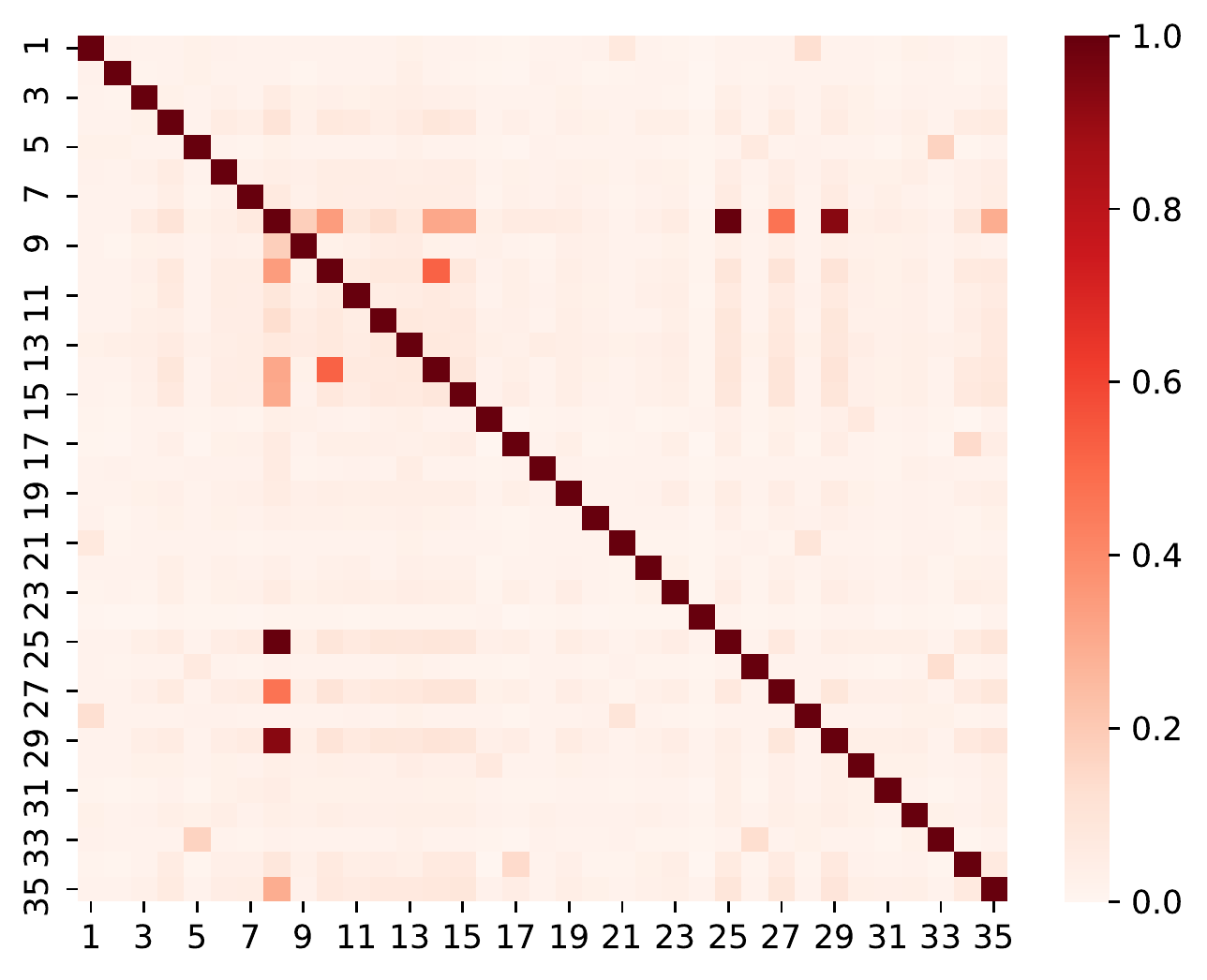}
                    }~~
                \subfigure[CME-DTW-d]{
                        \centering
                        \includegraphics[width=0.2\textwidth]{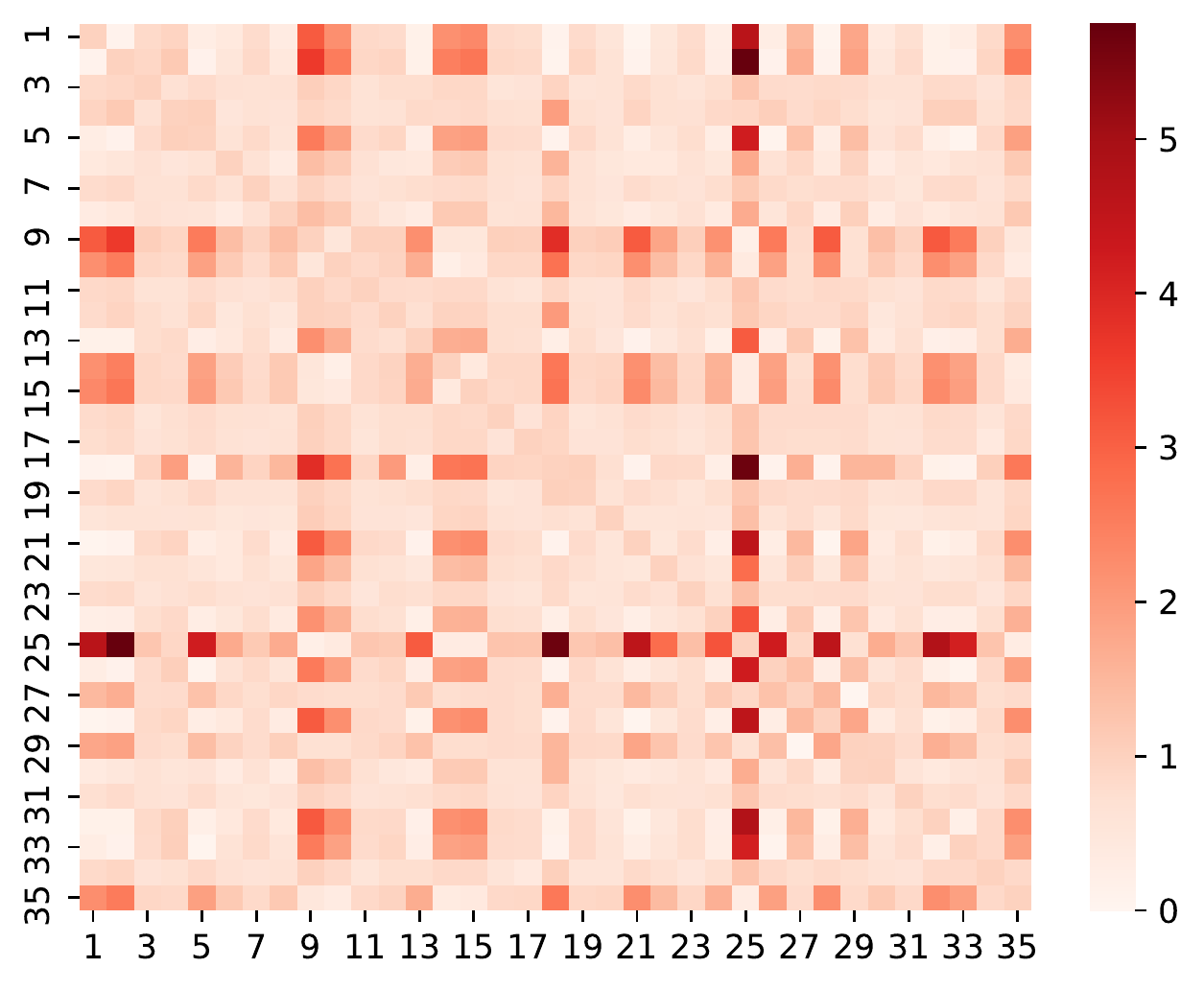}
                    }~~
                \subfigure[CME-GAK]{
                        \centering
                        \includegraphics[width=0.2\textwidth]{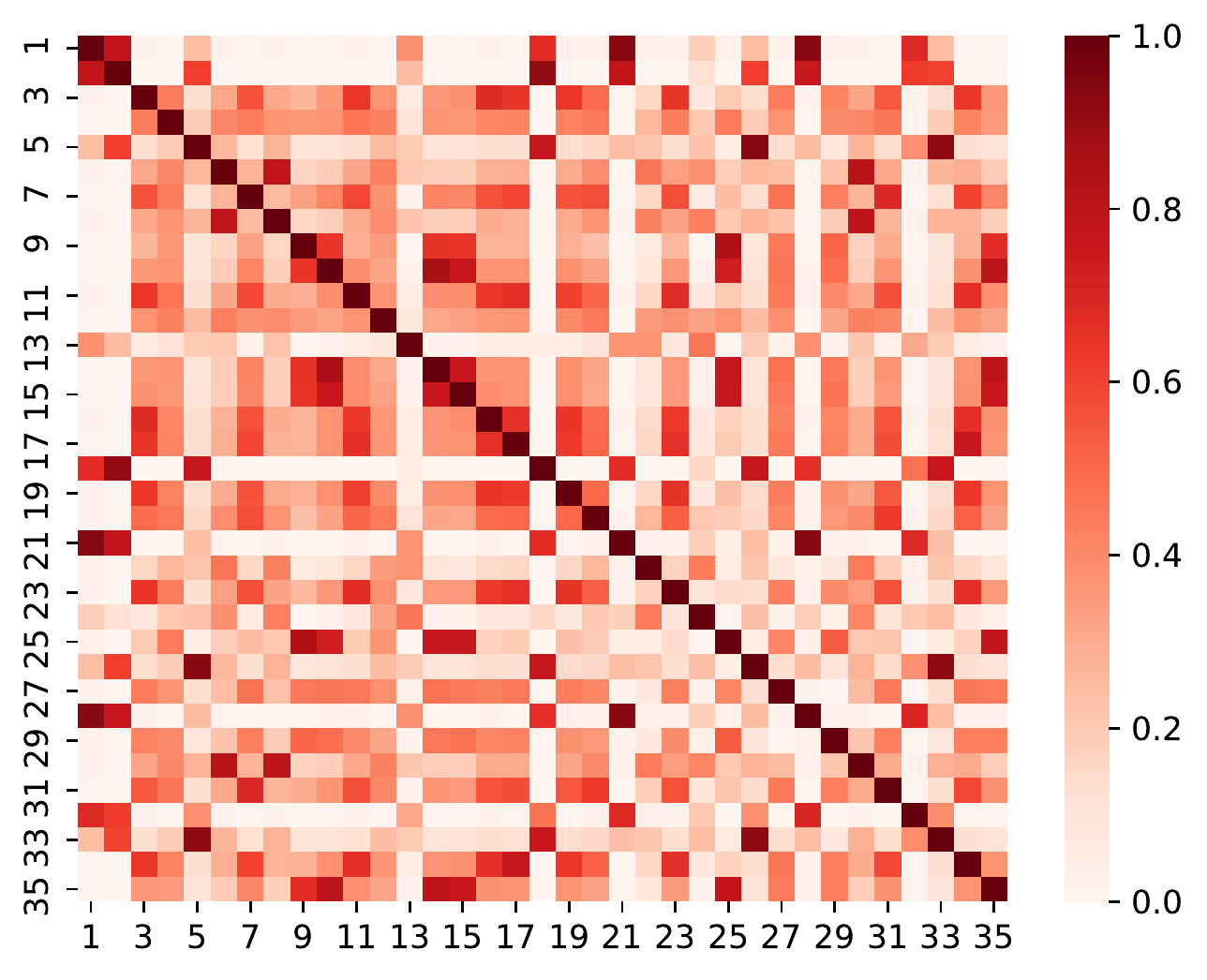}
                        }
            }
        }
    }
    \end{minipage}
    \\
    \begin{minipage}[t]{.88\textwidth}{
        \resizebox{\textwidth}{!}{
            \centering{~
                \subfigure[CME-PDTW(p=0.01)]{
                    \centering 
                    \includegraphics[width=0.2\textwidth]{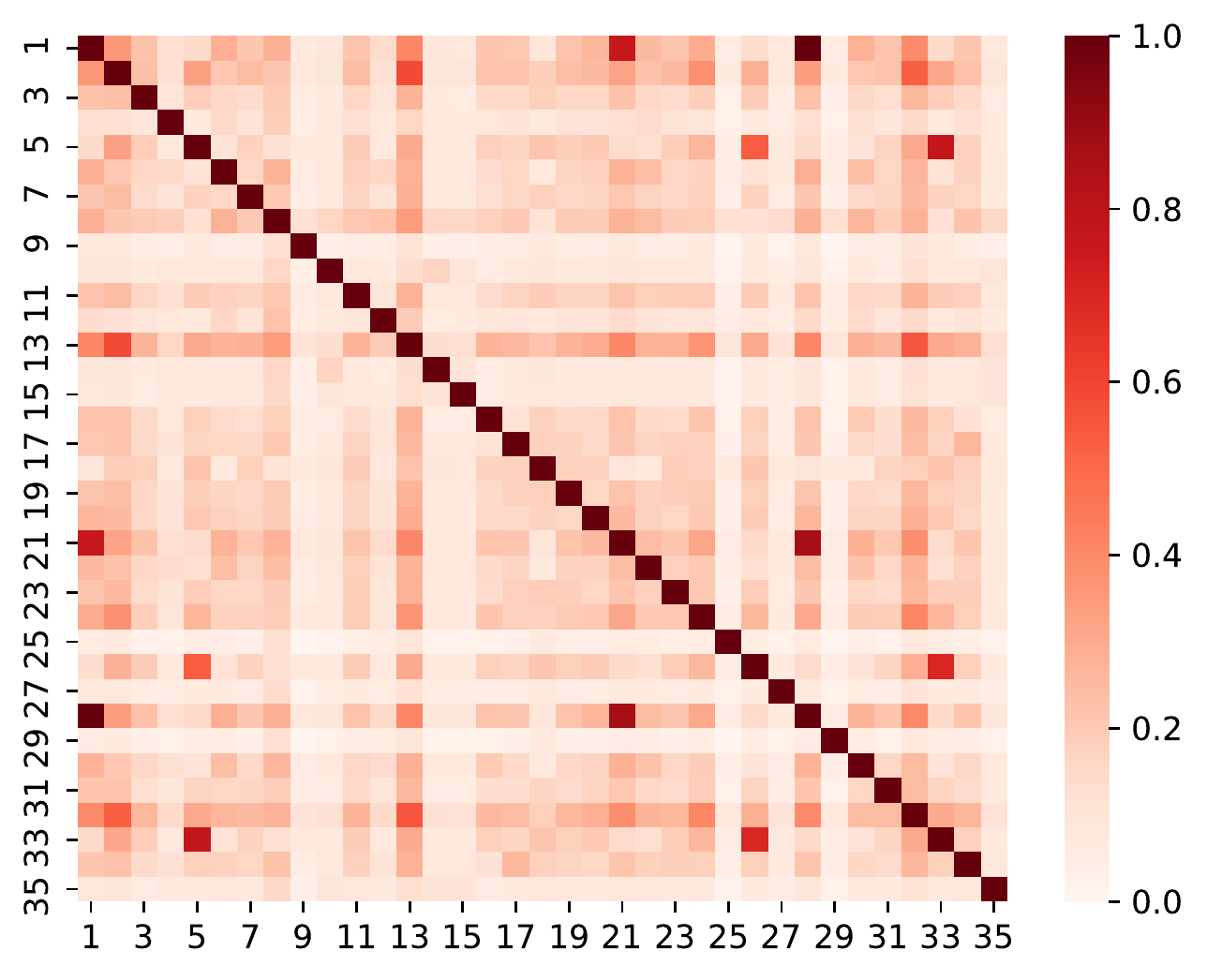}
                    }~~
                \subfigure[CME-PDTW(p=0.1)]{
                    \centering 
                    \includegraphics[width=0.2\textwidth]{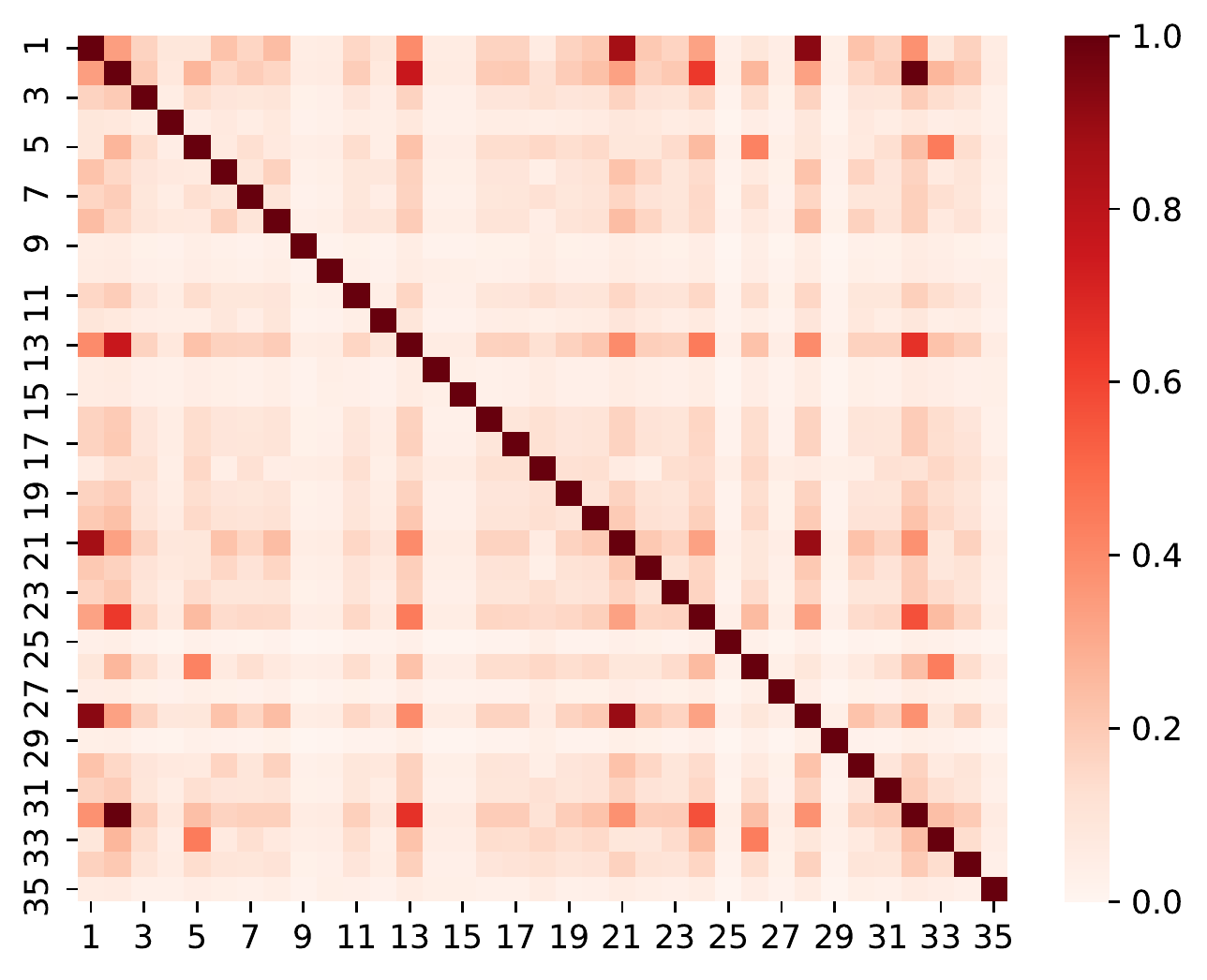}
                    }~~
                \subfigure[CME-PDTW(p=0.5)]{
                    \includegraphics[width=0.2\textwidth]{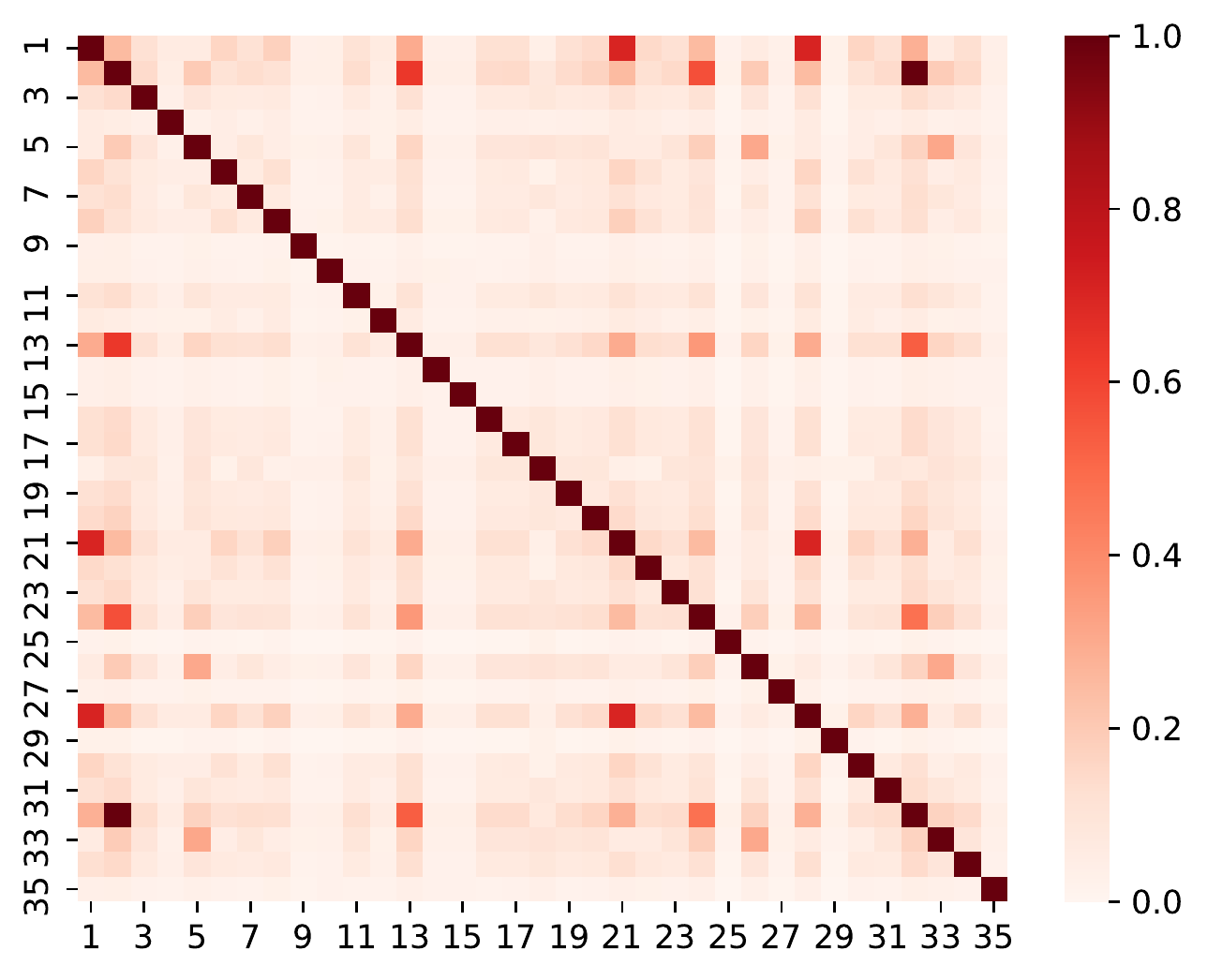}
                    }~~
                \subfigure[CME-PDTW(p=1.0)]{
                    \includegraphics[width=0.2\textwidth]{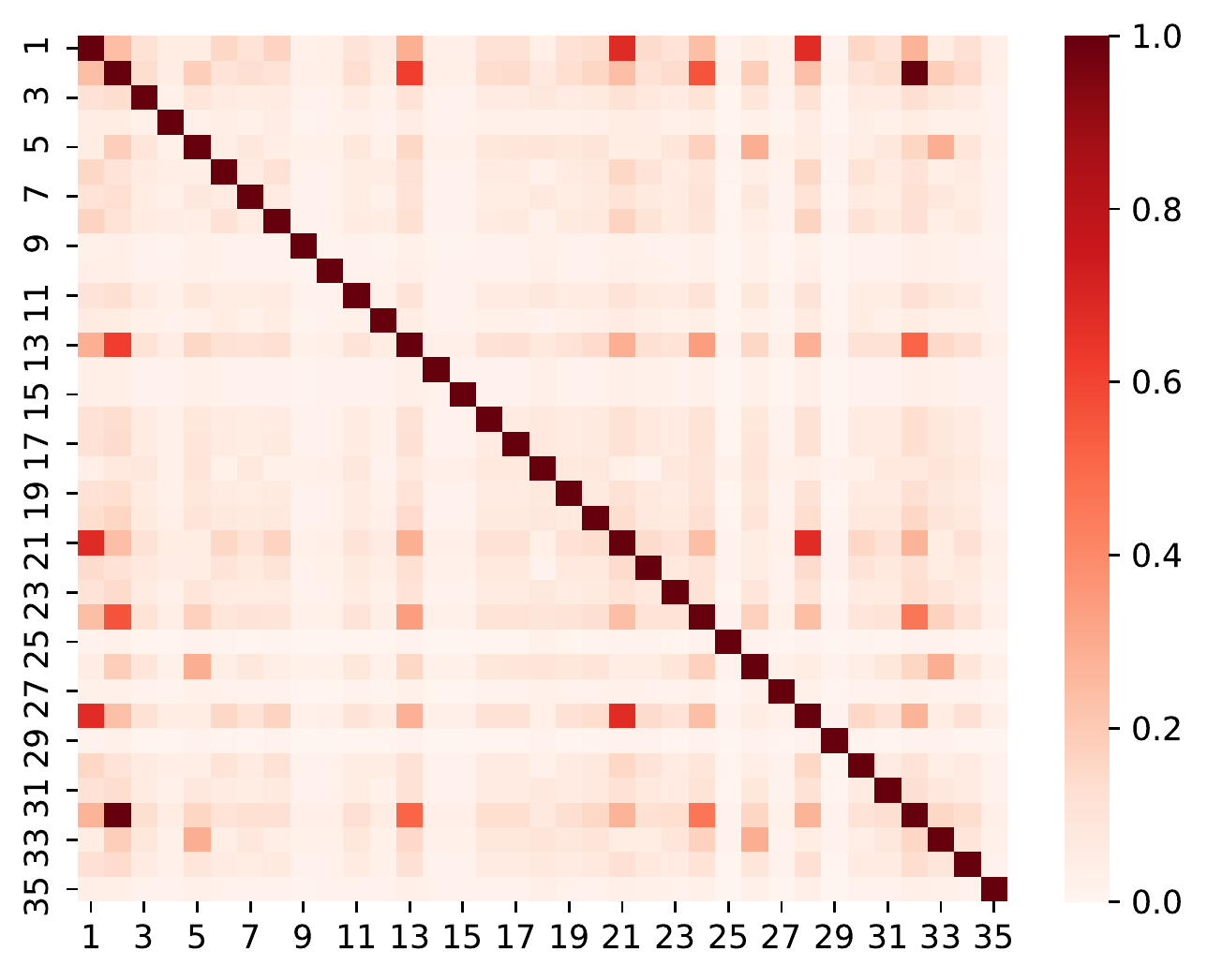}
                    }~~
                \subfigure[CME-PDTW(p=10)]{
                    \includegraphics[width=0.2\textwidth]{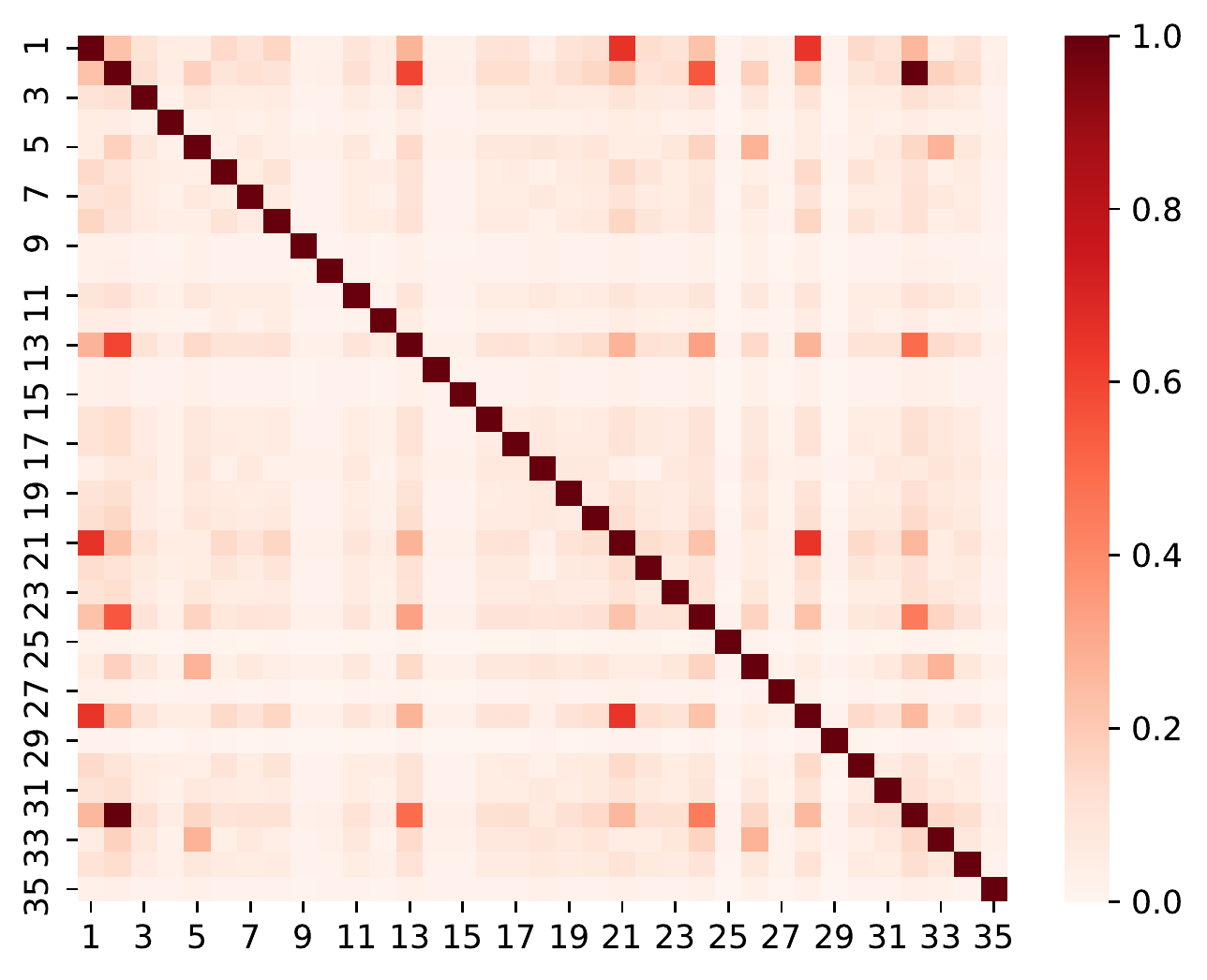}
                }
            }
        }
    }
    \end{minipage}
    }
   
    \centering
    \caption{Pairwise observation rate and correlation matrices extracted by different CME methods.}
    \label{fig:correlation_matrix}
    % \vspace*{-2pt}
\end{figure*}

We normalize the data to ensure that each input dimension has zero mean and unit variance. Following~\cite{che2018recurrent,song2018attend}, the missing values for all end-to-end models are treated as zero. We make sure all models have comparable numbers of parameters, that is, about 130~K parameters for PhysioNet, 40~K parameters for MIMIC-\uppercase\expandafter{\romannumeral3}, and 20~K parameters for human activity data, which are sufficient to make each model perform well at an acceptable computational cost. As the model complexity of LIFE is mainly dominated by the vector size $k$ of each individual feature, so the number of parameters requires $k\leq 6$ and we set $k=6$. For some parameters that have little impact on the complexity and performance of the model, we  set them to fixed values: the number of dense interpolated timestamps $F=3$ and the weight balancing factor $\alpha=1$. As for baseline methods, we used the codes provided by the authors for GRU-D based on keras\footnote{\url{https://github.com/keras-team/keras}} and BRITS based on PyTorch.  We implement SAnD, FG-LSTM based on PyTorch. The codes of MICE and ROCKET are available from scikit-learn\footnote{\url{https://github.com/scikit-learn/scikit-learn}} and sktime python package~\cite{sktime}. 
All deep models are trained by Adam optimizer~\cite{kingma2014adam} with learning rate $0.001$ and batch size 64. Training epochs for PhysioNet, MIMIC-\uppercase\expandafter{\romannumeral3}, and human activity data are 200, 400, and 100, respectively. The hyper-parameters are optimized by cross validation.  
We run the codes on Ubuntu 18.04 system with a single Nvidia TITAN XP graphic card and 32G memory. Our codes are available at \url{http://www.lamda.nju.edu.cn/code_LIFE.ashx}.

The baseline CME methods are based on Algorithm~\ref{alg:PDTW} with differences of missing-value processing in Step~\ref{alg:PDTW:interpolation} and of distance calculation in Step~\ref{alg:PDTW:dist}.

\begin{itemize}
    \item \textbf{Without Penalty}:
    \begin{itemize}
        \item \textbf{CME-Pearson} indicates the weighted absolute value of the pairwise Pearson Correlation Coefficient, where all missing values are excluded.
        \item \textbf{CME-DTW-i} and \textbf{CME-DTW-d} impute (linear interpolation) and drop the missing values, respectively, and then, conduct CME based on DTW.
        \item \textbf{CME-GAK}. Global Alignment Kernel (GAK)~\cite{cuturi2011fast} is a widely used kernel for time series. It drops missing values and calculate similarity via alignment and warping technics.
        % \item \textbf{CME-Pearson}. For each sample, calculate the pairwise Pearson Correlation Coefficient of all dimensions and any missing values are excluded. Then take the weighted average of the Correlation Coefficient matrix among all samples and the weight is the number of observations. Finally take the absolute value, we can get the correlation matrix.
        % \item \textbf{CME-DTW}. Apply Algorithm~\ref{alg:PDTW} but replace PDTW in Step~\ref{alg:PDTW:dist} by DTW. In fact, DTW is a special case of PDTW  when the penalty constant $p$ is 0.
        % \item \textbf{CME-GAK}. Global Alignment Kernel (GAK)~\cite{cuturi2011fast} is a wildly used positive definite kernel for time series. It can be used to MTS with missing values as it considers alignment and warping when assessing similarity~\cite{narayan2020survey}.
    \end{itemize}

    \item \textbf{With Penalty}:
    \begin{itemize}
        \item \textbf{CME-PDTW}. We apply the CME-PDTW algorithm and alter $p$ from $\{0.01, 0.1, 0.5, 1.0, 10\}$.
    \end{itemize}

\end{itemize}

\subsection{CME Results}
\label{sec:cme_results}
The first experiment is to evaluate the robustness and performance of the CME-PDTW algorithm and the baselines.
We conduct CME on the PhysioNet data set. Fig.~\ref{fig:correlation_matrix} displays the average pairwise observation rate (i.e., 1 minus missing rate) and the extracted correlation matrices.

From Fig.~\ref{fig:correlation_matrix}, we can see that the missing values will interfere with CME to a large extent. The correlation matrix of CME-DTW-i is inundated with fictitious correlations and far away from the ground truth.  The missing values bring serious fictitious correlations to it.  The correlation matrices obtained by CME-Pearson, CME-DTW-d, and CME-GAK are not stable. Most large values of the correlation matrices are corresponding to low observation rates.  However, CME-PDTW performs well and is not sensitive to the only hyper-parameter penalty coefficient $p$. As we can see, the correlation matrices tend to be stable with an increase of $p$. CME-PDTW can suppress the interference caused by missing values and obtain a credible and stable correlation matrix. We employ $p=0.5$ through the experiments.

\subsection{Performance Evaluation}
Following the previous work~\cite{cao2018brits,che2018recurrent}, we report the results of 5-fold cross validation and use the following evaluation criteria: area under ROC curve (AUC score) for mortality classification since it is a class imbalance task, mean absolute error (MAE) for length-of-stay prediction, and classification accuracy for human activity data.

\begin{table}[ht]
    % \vspace*{-3pt}
    \centering
    \caption{Classification performances on PhysioNet.}
    % \resizebox{.75\columnwidth}{!}{
        {\small
    \begin{tabular}{c c c c}
    \toprule
    &Models & AUC ($\pm$ std) & Paras ({\small $\sim\!$} K)\\
    \midrule
    \multirow{4}{20 pt}{Two-\\Steps} 
    &ROCKET-a & 0.8084 $\pm$ 0.018  & / \\
    &ROCKET-m & 0.8103 $\pm$ 0.015  & / \\
    &LSTM-a & 0.8091 $\pm$ 0.017 &   130.8 \\
    &LSTM-m  & 0.8046 $\pm$ 0.018 &  130.8 \\
    \midrule
    \multirow{5}{20 pt}{~End-\\ \ \ to-\\ \ End} 
    &GRU-D & 0.8379 $\pm$ 0.012  &  127.7 \\
    &BRITS & 0.8329 $\pm$ 0.008  &  129.0 \\
    &SAnD & 0.8011 $\pm$ 0.026  &  140.9 \\
    &FG-LSTM & 0.8193 $\pm$ 0.018 &   130.3 \\
    &\textbf{LIFE}  &  \textbf{ 0.8451 $\pm$ 0.012 } & \textbf{129.3}\\
    \bottomrule
    \end{tabular}}
    % }
    \label{tbl:PhysioNet}
\end{table}

The results on the PhysioNet data set, as shown in Table~\ref{tbl:PhysioNet}, indicate that the LIFE model outperforms the others. Note that the two steps methods perform significantly worse than the SOTA end-to-end methods, which is consistent with previous research~\cite{che2018recurrent,cao2018brits,tang2020joint}. So in the following experiments, we only focus on end-to-end models.

\begin{table}[ht]
    % \vspace*{-3pt}
    \centering
    \caption{Model Comparison on MIMIC-\uppercase\expandafter{\romannumeral3} data set.}
    % \resizebox{.75\columnwidth}{!}{
        {\small
    \begin{tabular}{c c c c}
    \toprule
    \multirow{2}{*}{Models}    & Classification & Forecasting & Paras\\
    & AUC ($\pm$ std) & MAE ($\pm$ std)  & ({\small $\sim\!$} K) \\
    \midrule
    GRU-D & 0.8457 $\pm$ 0.005  &  2.874 $\pm$ 0.203 &   40.1 \\
    BRITS & 0.8435 $\pm$ 0.008  & 2.631  $\pm$ 0.184 &   41.5 \\
    SAnD & 0.8244 $\pm$ 0.018  & 3.012 $\pm$ 0.242 &   42.7 \\
    FG-LSTM & 0.8332 $\pm$ 0.015 &  2.781 $\pm$ 0.186 &   41.2 \\
    \textbf{LIFE}  &  \textbf{ 0.8629 $\pm$ 0.007 } & \textbf{2.556 $\pm$ 0.146} & \textbf{ 40.3 } \\
    \bottomrule
    \end{tabular}}
    % }
    \label{tbl:mimic3}
\end{table}

Table~\ref{tbl:mimic3} shows the performances of different end-to-end models. In both classification and forecasting tasks, LIFE achieves the best performance.

\begin{figure}[t]
    \centering
    \includegraphics[width=.95\columnwidth]{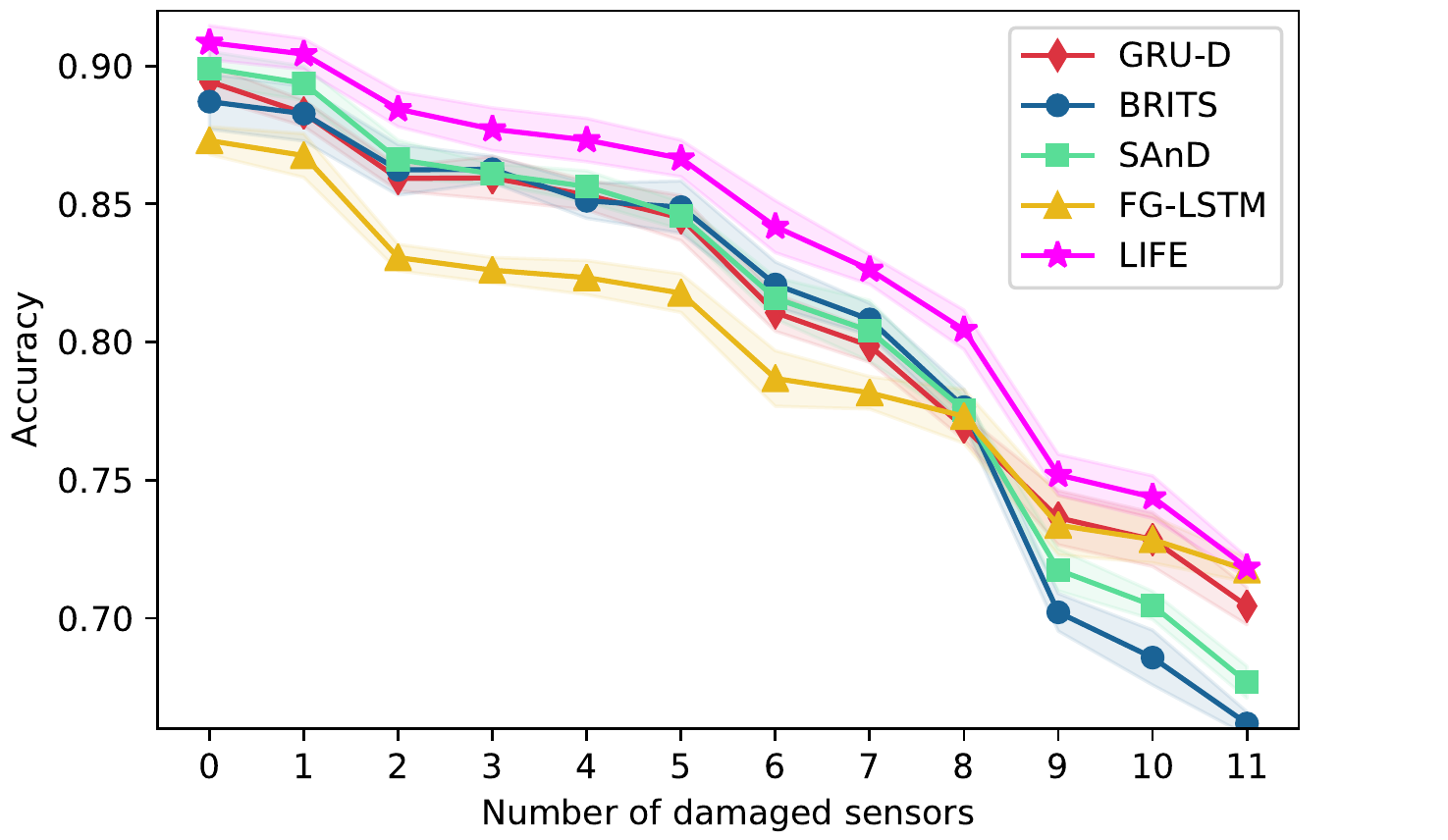}
    \caption{Performances on human activity data with different numbers of damaged sensors. Shaded regions is the standard deviation over five trials.}
    \label{fig:human_activity}
    % \vspace*{-15pt}
\end{figure}

We further investigate the performances of models with different missing rates. The mean and standard deviation of accuracy on the human activity data sets are shown in Fig.~\ref{fig:human_activity}. The more damaged sensors are, the higher the missing rate is. We can see that LIFE achieves the best results and significantly outperforms the others on most of the missing rates. When the missing rate is high, it can suppress the interference of missing values and generate as reliable features as possible.  When there are few missing values, it also explores correlations effectively and maintains good performance. An interesting aspect of this result is FG-LSTM. It shows that handling each input dimension separately can bring benefits when the missing rate is high, but it can also damage the performance when the missing rate is low. Therefore, it is important to not only suppress the interaction of missing values but also utilize dimensional correlations, as LIFE does.
\subsection{Individual Features Construction Comparison}
In order to explore whether the individual feature construction does bring benefits, we keep the hyper-parameters of  LIFE  consistent and change the correlation matrices: 
\begin{itemize}
    \item \textbf{Ones} denotes the Full-1 matrix, which leads LIFE to fuse all input dimensions, just as most existing models do. 
    \item \textbf{Rand} is a random matrix sampling from $[0,1]$. 
    \item \textbf{Diag} is a diagonal-1 matrix, which makes LIFE build feature for each input dimension independently. 
    \item \textbf{CME-Pearson}, \textbf{CME-DTW-i}, \textbf{CME-DTW-d}, \textbf{CME-GAK}, and \textbf{CME-PDTW} are correlation matrices extracted via Algorithm~\ref{alg:PDTW} with different distances.
\end{itemize}

\begin{table}[ht]
    % \vspace*{-3pt}
    \centering
    \caption{Classification performances of LIFE with different correlation matrices on PhysioNet.}
    \resizebox{.87\columnwidth}{!}{
        {\small
        \begin{tabular}{c c c}
            \toprule
            Correlation Matrix  &AUC ($\pm$ std) & Paras ({\small $\sim\!$} K)\\
            \midrule
            Ones & 0.8204 $\pm$ 0.010 &  129.3 \\ 
            Rand& 0.8251 $\pm$ 0.009 &  129.3  \\
            Diag& 0.8355 $\pm$  0.012 &  129.3 \\
            CME-Pearson& 0.8336 $\pm$ 0.007&  129.3  \\
            CME-DTW-i& 0.8232 $\pm$ 0.010 &  129.3  \\
            CME-DTW-d& 0.8346 $\pm$ 0.009 &  129.3  \\
            CME-GAK& 0.8374 $\pm$  0.012 &  129.3 \\
            \textbf{CME-PDTW}& \textbf{ 0.8451 $\pm$ 0.012 } & \textbf{129.3}\\
            \bottomrule
            \end{tabular}
        }
        }
    \label{tbl:correlation_matrix}
\end{table}

The results of LIFE on the PhysioNet data set are shown in Table~\ref{tbl:correlation_matrix}. 
LIFE with the correlation matrix extracted by CME-PDTW achieves the highest AUC score, which shows the effectiveness of building individual features and the CME-PDTW algorithm. On the contrary, Ones performs significantly worse than CME-PDTW, verifying that merging all input dimensions is not acceptable since it will damage the classification performance. Diag, CME-Pearson, CME-DTW-d, and CME-GAK also achieve competitive AUCs, verifying that building individual features can bring benefits. The poor performances of Rand and CME-DTW-i show the wrong correlations can damage the downstream prediction performance.

\section{Conclusion}
\label{sec:6}
In this paper, we proposed a novel framework LIFE, which provides a new paradigm for MTS prediction with missing values. For each input dimension, LIFE utilizes credible and correlated dimensions to build individual features. So LIFE is able to not only suppress the interference of missing values but also generate reliable and effective features for prediction. Besides, we also propose the idea of penalizing missing values for CME and provide two algorithms as concrete implementations. Experiments conducted on three real-world data sets show that LIFE outperforms the existing SOTA models and our CME algorithms can extract credible and stable correlations.
% \\
% \\
% \textbf{Acknowledgment:} This research was supported by the National Key Research and Development Program of China No.2020AAA0109400. 
% The corresponding author is Yuan Jiang.

\section*{Acknowledgment}

This research was supported by the National Key Research and Development Program of China (2020AAA0109400). The corresponding author for this work is Yuan Jiang.

\appendix
\renewcommand\appendixname{Appendix of LIFE}

\subsection{CME with Penalty}
Our approach that penalizing missing information CME with missing values can be implemented with various distances. The challenge lies in designing the penalty term. We have introduced CME-PDTW in Section~\ref{sec:4}. Here we give an alternative implementation with Optimal Transport (OT)~\cite{villani2008optimal} for CME. We first introduce the information of OT and then apply Penalty Optimal Transport (POT) to time series.

\subsubsection{Optimal Transport (OT)}
OT (a.k.a. Wasserstein Distance) is a powerful tool to compare probabilities or histograms. Given a transport cost function (a.k.a. ground metric), OT finds the transportation plan that minimizes the total expected cost. We denote two probability distributions of $L$ elements as $\left(\mathcal{A} \mid \boldsymbol {u}\right)=\left\{\left(a_{i} \mid u_i \right)\right\}_{i=1}^L$ and $\left( \mathcal{B} \mid \boldsymbol {v}\right)=\{(b_{j} \mid v_j)\}_{j=1}^L$, respectively, where $a_i$ ($b_i$) is the $i$-th value in the  distribution and $u_i (v_i) \in [0,1]$ is its corresponding probability value. $\mathbf{\Omega} \in \mathbb{R}_+^{L \times L}$ indicates the cost matrix, where ${\Omega}_{ij}$ is the cost between $a_i$ and $b_j$. The OT distance is calculated as:
\[
\text{OT}(\mathcal{A}, \mathcal{B}) =\min _{\mathbf{P}  \in \mathrm{U}(\mathcal{A}, \mathcal{B})} \sum_{i, j=1}^{L} \mathrm{P}_{ij} \mathrm{\Omega}_{ij}
\]
where $\mathrm{U}(\mathcal{A}, \mathcal{B}) \overset{\underset{\mathrm{def}}{}}{=} \left\{\mathbf{P} \in [0,1]^{L \times L} \mid \mathbf{P} \  \boldsymbol 1= \boldsymbol {u}, \mathbf{P}^{\top} \boldsymbol 1=\boldsymbol {v}\right\} $ denotes the joint probability measures such that the first marginal is $\boldsymbol {u}$ and the second marginal is $\boldsymbol {v}$. The computational cost of OT can be significantly reduced by entropy regularization~\cite{cuturi2013sinkhorn}.

\begin{algorithm}[t]
	\caption{POT distance}
	\label{alg:POT}
	\begin{algorithmic}[1]
		\Require 
		Univariate time series $\mathcal{A}=[a_i]_{i=1}^T$, $\mathcal{B}=[b_j]_{j=1}^T$, masking sequences $[m^a_i]_{i=1}^T$, $[m^b_j]_{j=1}^T$, time interval sequences $[\delta^a_i]_{i=1}^T$, $[\delta^b_j]_{j=1}^T$, and penalty coefficient $p$.
		\Ensure
		POT distance $dist$.
		\State $\boldsymbol t = \operatorname{zscore}\left(\operatorname{linspace}(1,T,T)\right)$ \Comment{normalize time index}
		\State $\mathbf{\Omega} = \operatorname{zeros}(T,T)$ \Comment{cost matrix}
		\label{alg:POT:Q}
		\For{$i = 1 \to T$}
		\For{$j = 1 \to T$}
		\State ${\Omega}_{ij} = (a_i-b_j)^2 + \beta (t_i-t_j)^2 + \phi(i,j) $ 
		\EndFor
		\EndFor
		\label{alg:POT:Q_end}
		\State $\boldsymbol u= \boldsymbol v=\operatorname{ones}(T)/T$
		\State $dist = \operatorname{OTSolver}(\boldsymbol u, \boldsymbol v,\mathbf{\Omega})$
	\end{algorithmic}
\end{algorithm} 
\subsubsection{Penalty Optimal Transport (POT)}
When applying OT to time series, besides the distance of observed values, it is also necessary to consider the temporal dimension. We adopt the Time Adaptive Optimal Transport (TAOT)~\cite{zhang2020time}, which defines the transport cost as a combination of observed value distances and temporal shifts. 
For the sake of brevity, we assume all time series are sampled uniformly and have the same length.
In TAOT, the one dimensional data point, for example $a_i$, is extended by another time dimension as $(a_i,t_i)$.
TAOT assumes all values of a time series share an equal probability. As a result, the input time series become
\begin{equation}\begin{array}{l}
\left(\mathcal{A} \mid \boldsymbol {u}\right)=\left\{\left(a_{i},t_i  \mid 1/L \right)\right\}_{i=1}^L, \\
\left(\mathcal{B} \mid \boldsymbol {v}\right)=\{(b_{j},t_j  \mid 1/L)\}_{j=1}^L.
\end{array}\end{equation}
Then the cost ${\Omega}_{ij}$ between $(a_i,t_i)$ and $(b_j, t_j)$ is defined as: 
\begin{equation}
    \label{eqn:cost}
    \mathrm{\Omega}_{ij} = (a_i-b_j)^2 + \beta (t_i-t_j)^2,
\end{equation} where $\beta$ is the time distance weight to balance two parts.
Then we can obtain POT by designing a penalty term to TAOT. To penalize missing values, as in PDTW, we add a penalty term $\phi$ and calculate $\mathrm{\Omega}^\prime_{ij}$ according to:
\[{\Omega}^\prime_{ij} = (a_i-b_j)^2 + \beta (t_i-t_j)^2 +  \phi \left(i, j\right).
\label{eqn:pot_Q}
\]
The penalty term $\phi$ is the same definition as that in PDTW. It imposes more punishment on the consecutive missing values:
\[
    \phi \left(i, j\right) = p~\left({ \delta^a_{i}}\left(1-\rm m^a_i\right)+ {\delta^b_{j}}\left(1-{\rm m^b_j}\right)\right),
\] 
where $p$ indicates the penalty coefficient, $m^a_i (m^b_j)$ denotes whether $a_i (b_j)$ is missing, and $\delta^a_i (\delta^b_j)$ is the the time gap from the timestamp of last observation to current timestamp. When $p=0$, TAOT becomes a special case of POT.

Algorithm~\ref{alg:POT} introduces the calculation process of POT distance between two univariate time series. Since we have computed the cost matrix $\mathbf{\Omega}$ (Lines~\ref{alg:POT:Q}-\ref{alg:POT:Q_end}), we can obtain the OT distance through an OT solver. Furthermore, to use POT distance for CME, we can apply CME-PDTW (Algorithm \ref{alg:PDTW}) and replace PDTW in Line~\ref{alg:PDTW:dist} by POT. 
\begin{figure*}[ht]
    \centering
    \vspace*{-5pt}
	{\small
		\begin{minipage}[t]{0.9\linewidth}{
				\centering{
					~~~~
					\subfigure[p=0]{
						\centering
						\includegraphics[width=0.17\linewidth]{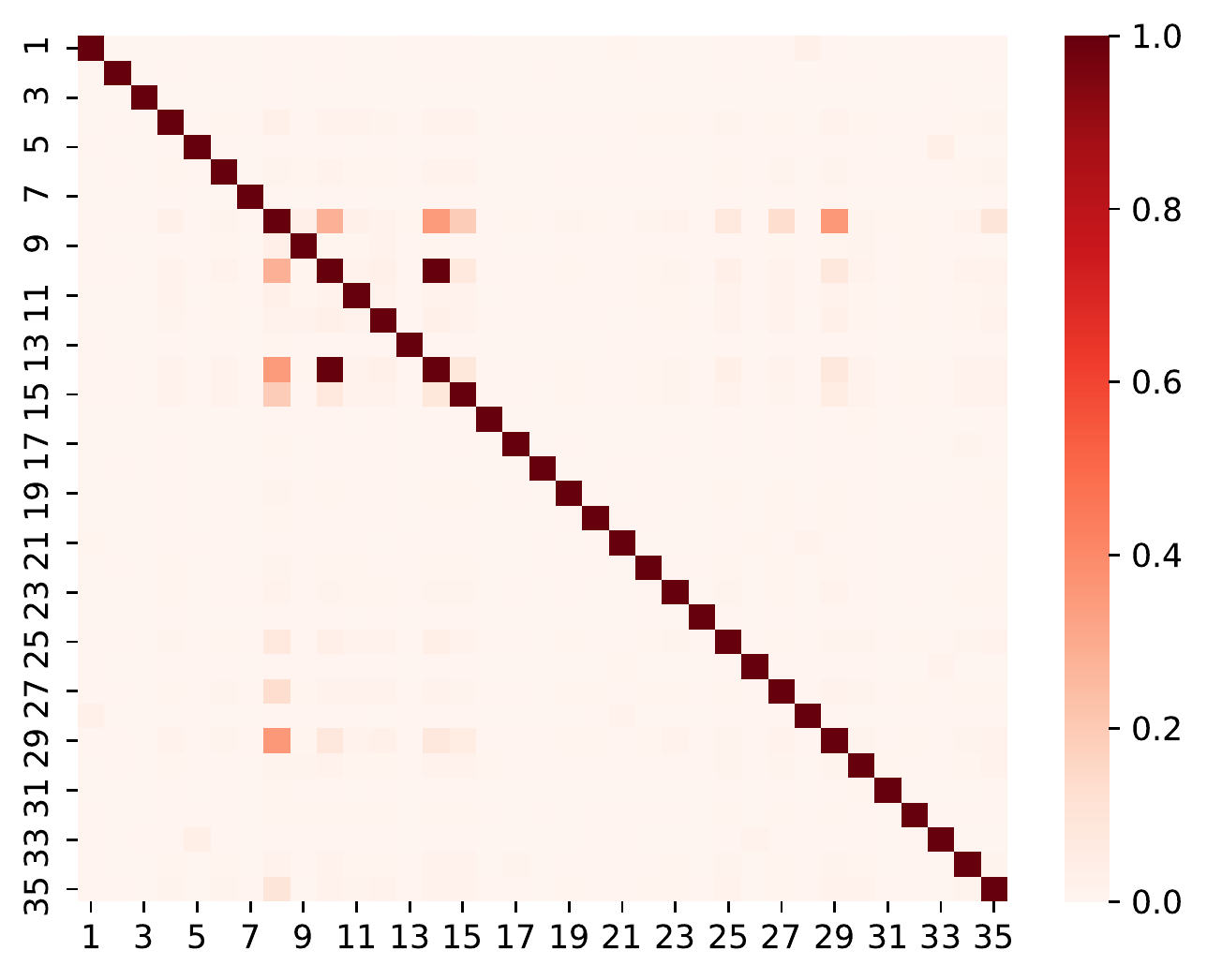}
					}~~~~~~~~
					\subfigure[p=0.001]{
						\centering
						\includegraphics[width=0.17\linewidth]{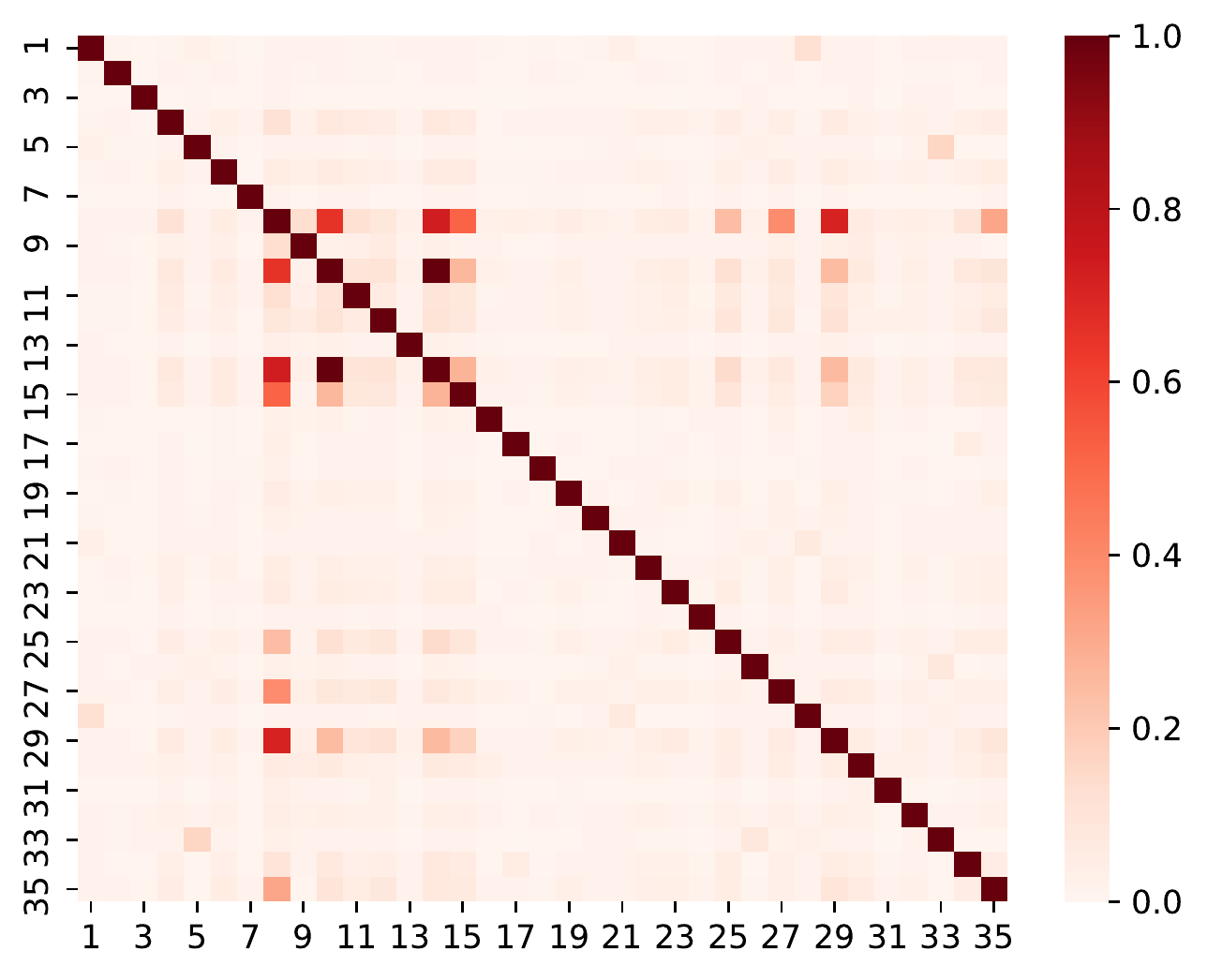}
					}~~~~~~~~
					\subfigure[p=0.01]{
						\centering
						\includegraphics[width=0.17\linewidth]{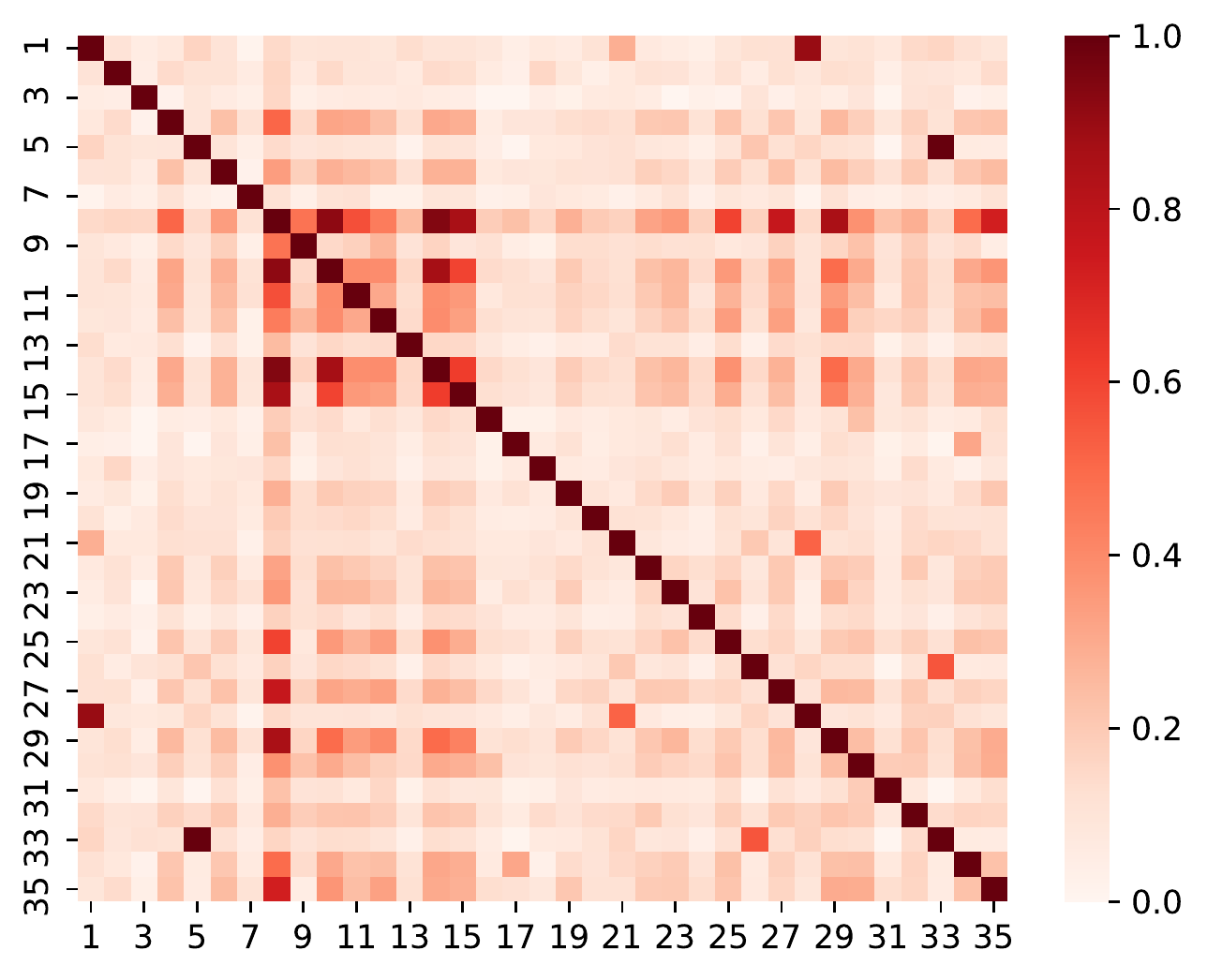}
					}~~~~~~~~
					\subfigure[p=0.1]{
						\centering
						\includegraphics[width=0.17\linewidth]{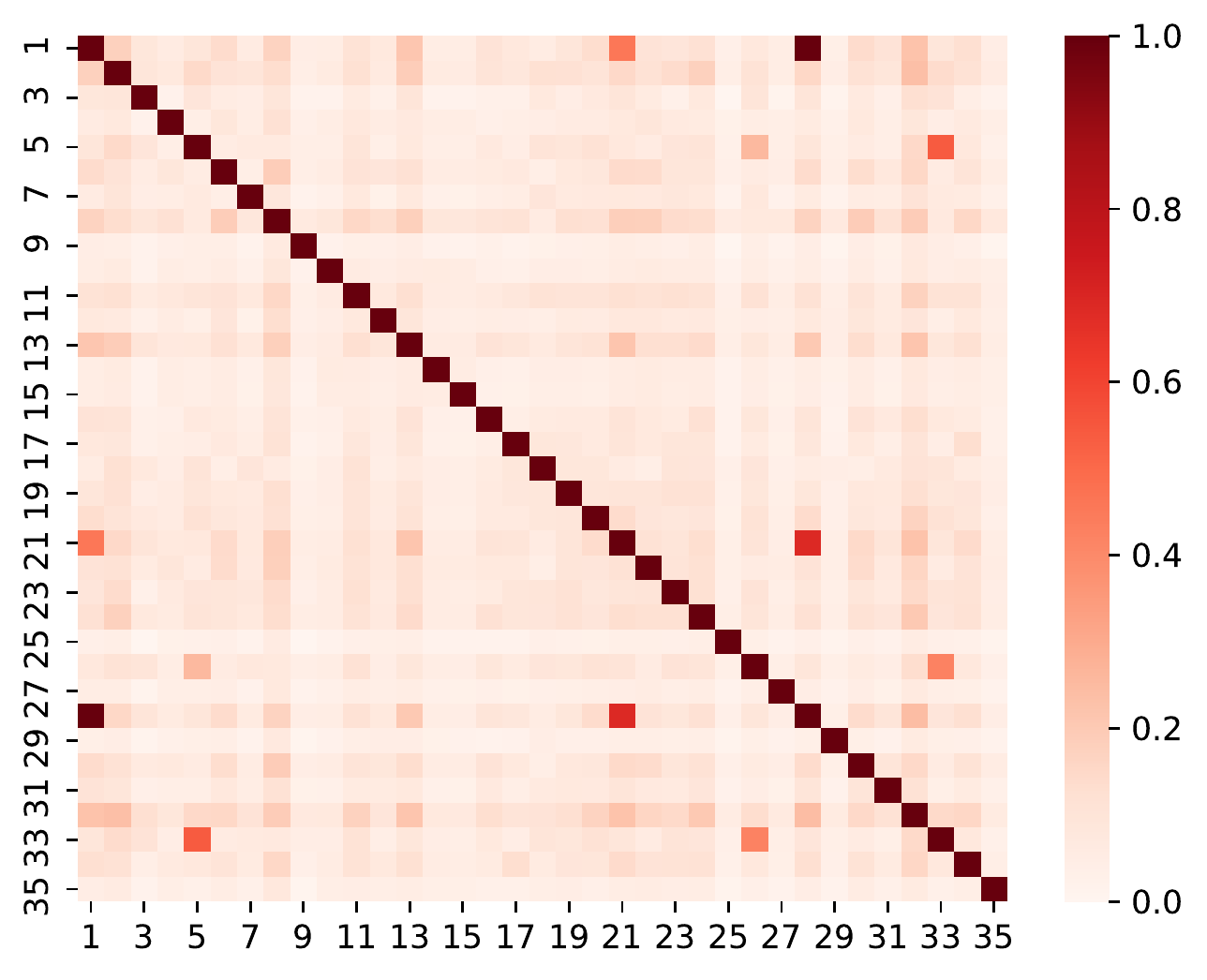}
					} 
				}
			}
		\end{minipage}
        \\
        \vspace{-4pt}
		\begin{minipage}[t]{0.9\linewidth}{
				\centering{
					\subfigure[p=0.5]{
						\centering 
						\includegraphics[width=0.17\linewidth]{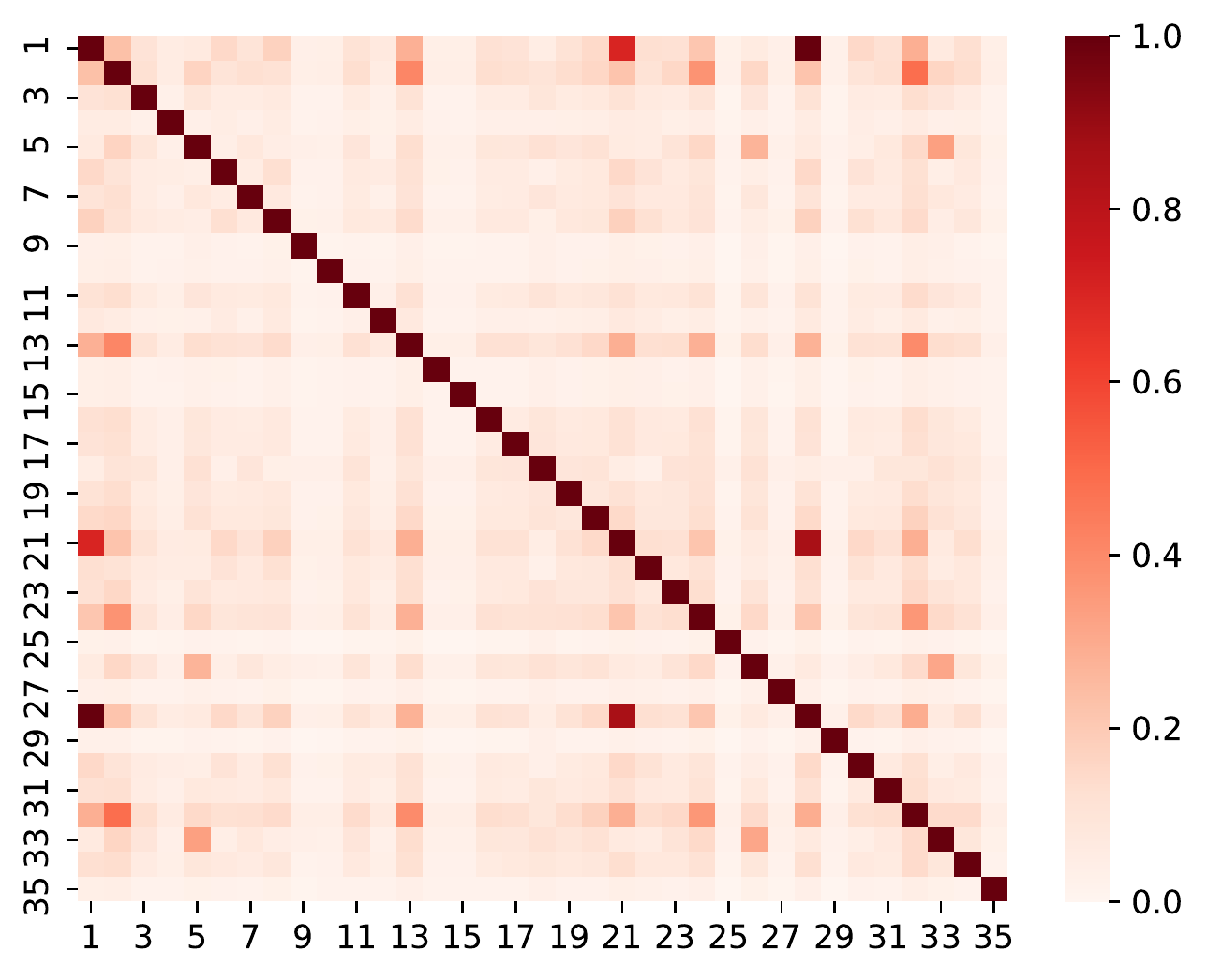}
					}~~~~~~~~
					\subfigure[p=1.0]{
						\includegraphics[width=0.17\linewidth]{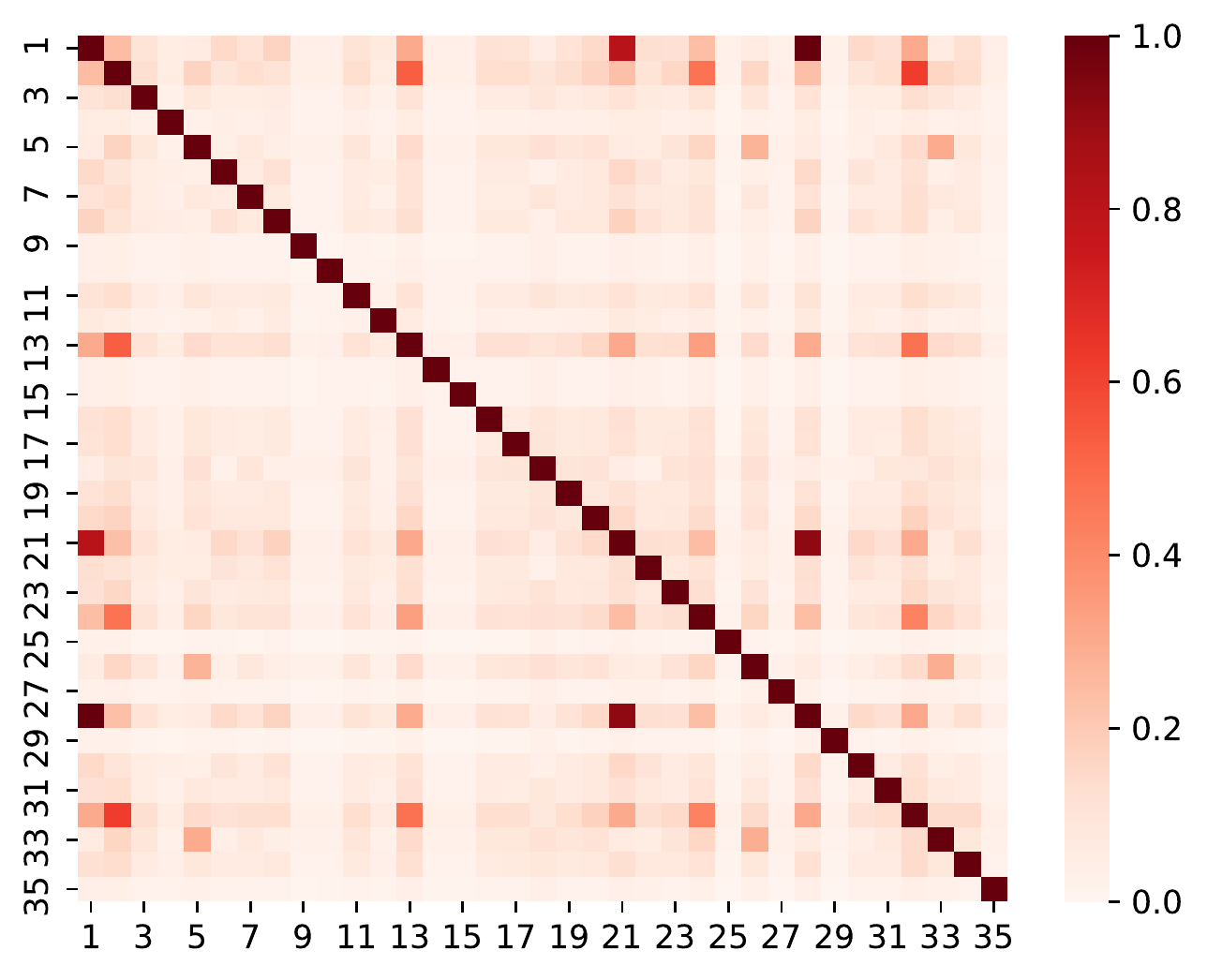}
					}~~~~~~~~
					\subfigure[p=10]{
						\includegraphics[width=0.17\linewidth]{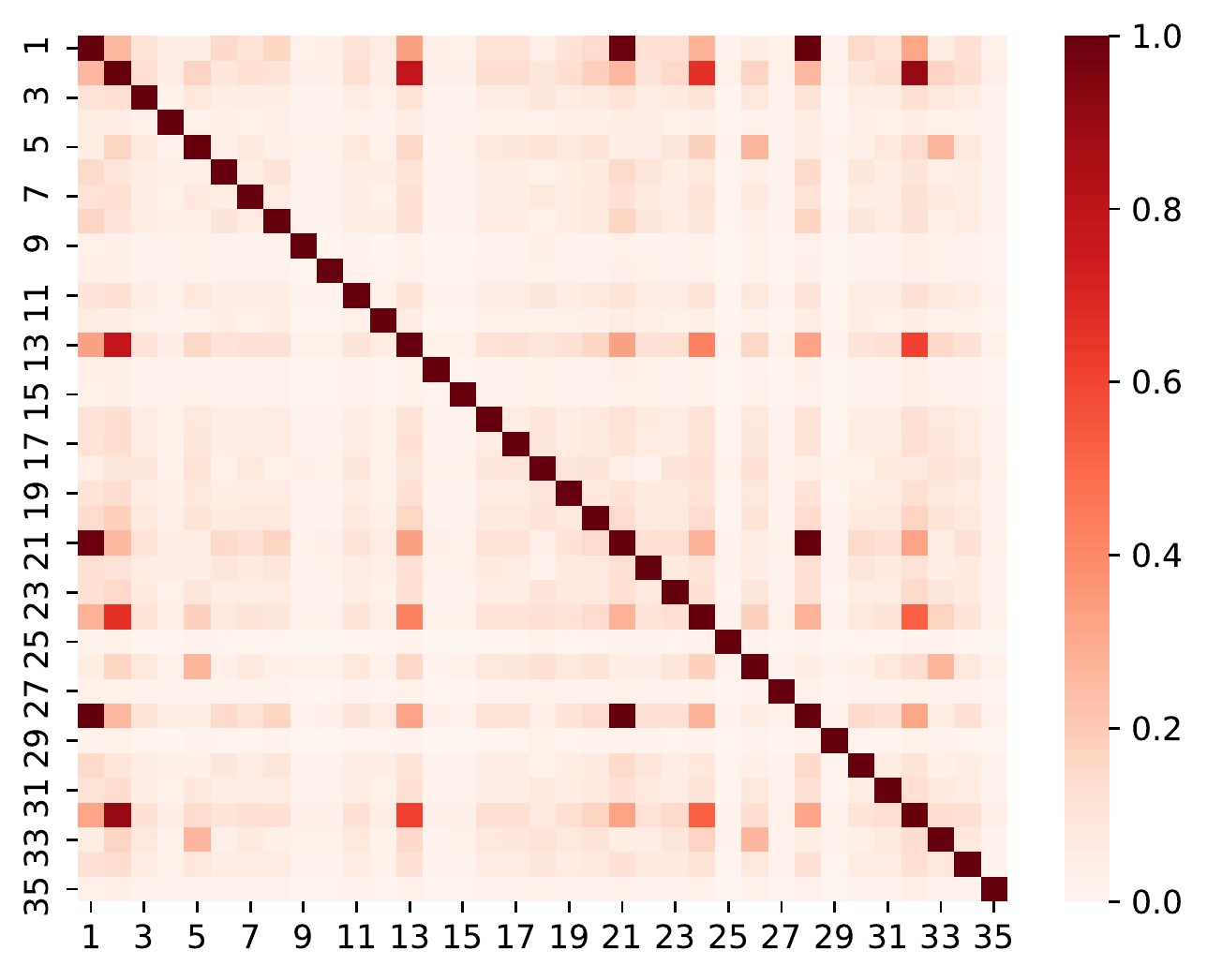}
					}~~~~~~~~
					\subfigure[p=100]{
						\includegraphics[width=0.17\linewidth]{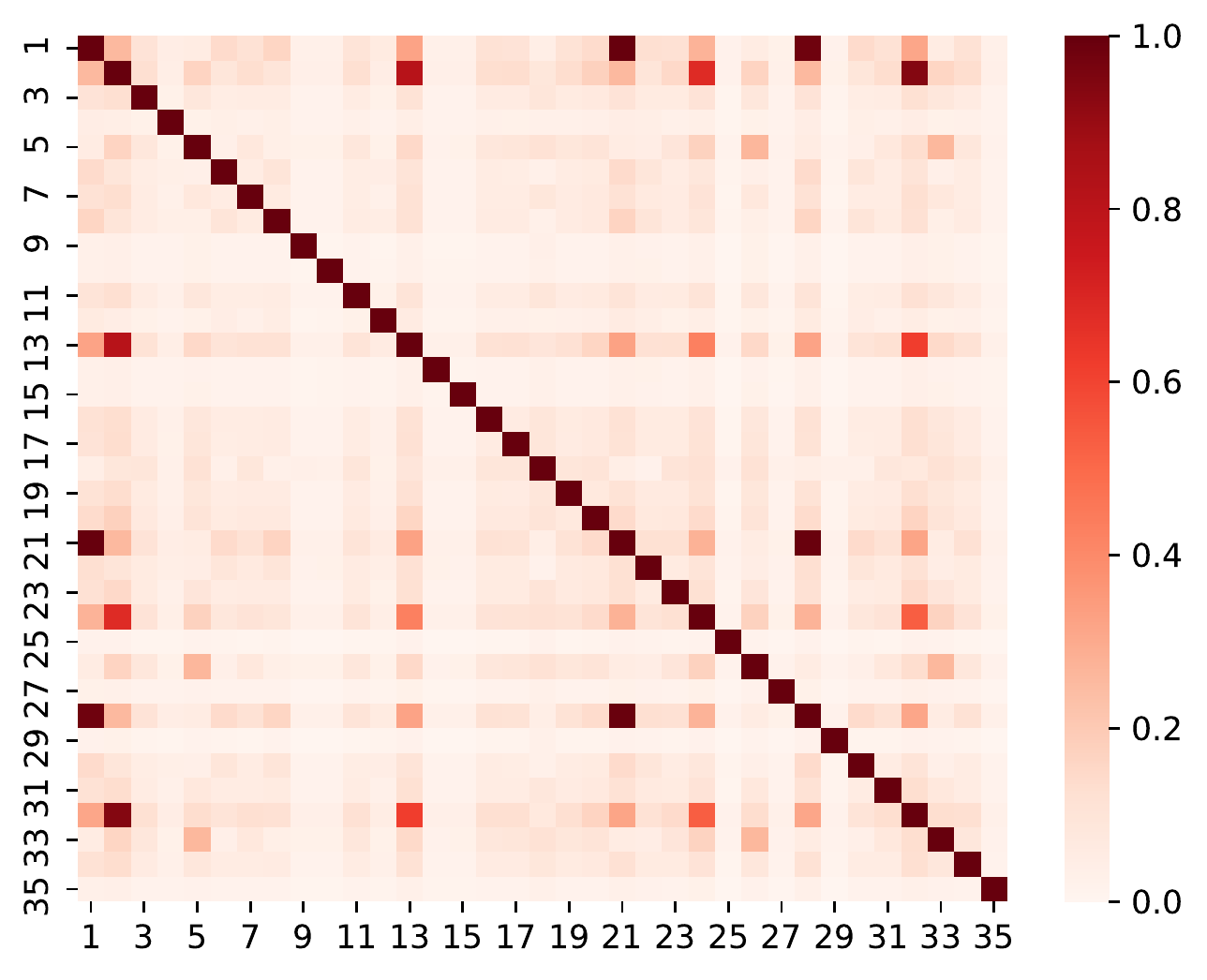}
					}
				}
				
			}
		\end{minipage}
	}
	
	\centering
	\caption{Correlation matrices extracted by CME-POT with different penalty constant $p$.}
	\label{fig:correlation_matrix_POT}
	% \vspace*{-10pt}
\end{figure*}

\subsubsection{Results of CME-POT}
The experiments of CME-POT are conducted on the PhysioNet data set. We fix the time distance weight $\beta$ to 1 and alter penalty constant $p$ from $\{0, 0.001, 0.01, 0.1, 0.5, 1.0, 10, 100\}$.  The extracted correlation matrices are shown in Fig.~\ref{fig:correlation_matrix_POT}. With the increase of $p$, the correlation matrices become stable, which is consistent with the results of CME-PDTW. Besides, the stable correlation matrices of CME-POT are very similar to those of CME-PDTW, which shows that the idea of penalizing missing values is a general method to seek credible and stable dimensional correlations for CME with missing values.

% \newpage

{\footnotesize
\bibliographystyle{IEEEtranS}
\bibliography{ref}
}

\end{document}